%% file: main.tex
\documentclass[10pt,twocolumn,letterpaper]{article}

\usepackage[pagenumbers]{cvpr}           

\usepackage{float} 
\usepackage{graphicx}
\usepackage{amsmath}
\usepackage{amssymb}
\usepackage{booktabs}
\usepackage{enumitem}
\usepackage{cuted}
\usepackage{multirow}
\usepackage{ragged2e}
\newcommand{\paravspace}{\vspace{-8pt}}

\definecolor{cvprblue}{rgb}{0.21,0.49,0.74}
\usepackage[pagebackref,breaklinks,colorlinks,allcolors=cvprblue]{hyperref}

\def\paperID{4017} 
\def\confName{CVPR}
\def\confYear{2026}

\title{\!\!HiSpatial: Taming Hierarchical 3D Spatial Understanding in \\ Vision-Language Models\!\!}

\author{
Huizhi Liang$^{1,2,*\ddagger}$ \quad
Yichao Shen$^{3,2,*\ddagger}$ \quad
Yu Deng$^{2}$ \quad
Sicheng Xu$^{2}$ \quad
Zhiyuan Feng$^{1,2,\ddagger}$ \\
Tong Zhang$^{4}$ \quad
Yaobo Liang$^{2}$ \quad
Jiaolong Yang$^{2}$ \\
$^{1}$Tsinghua University \quad
$^{2}$Microsoft Research Asia \quad
$^{3}$Xi'an Jiaotong University \\
$^{4}$ University of the Chinese Academy of Sciences }

\begin{document}

\maketitle
\begingroup
\renewcommand\thefootnote{}\footnotetext{
\hspace*{-1.8mm}\begin{tabular}{@{}l@{}}
$^*$Equal contribution\\
$^\ddagger$Work done during internship at Microsoft Research Asia\\

\end{tabular}
}
\endgroup

\begin{strip}
\vspace*{-65pt} 

\centering
{\large \url{https://microsoft.github.io/HiSpatial/}} 

\vspace{15pt}

\includegraphics[width=1\textwidth]{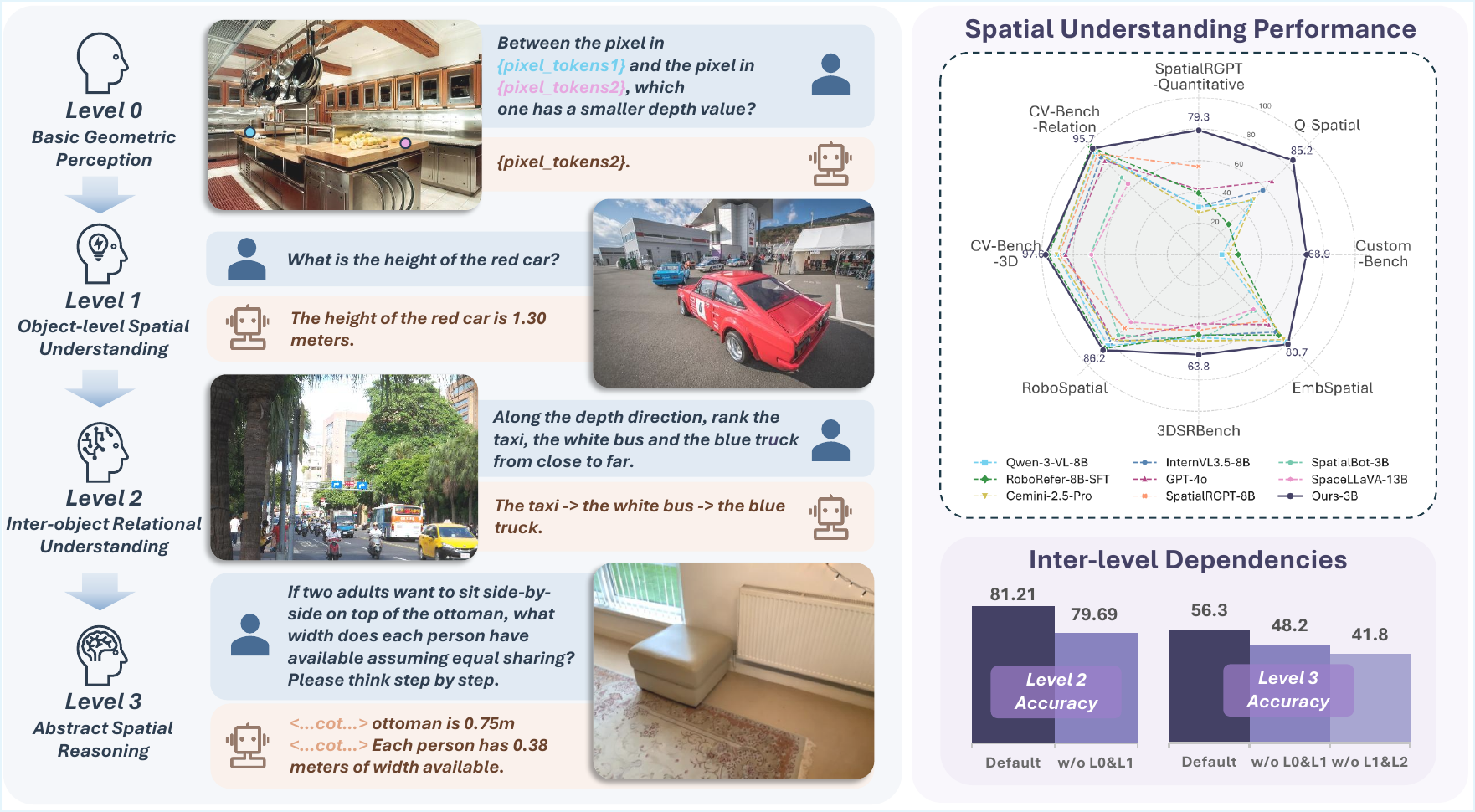}
\vspace{-8pt}
\captionsetup{type=figure,font=small,position=top}
\caption{Trained on our large-scale spatial VQA data, our model develops hierarchical 3D spatial intelligence from geometric perception to abstract reasoning (left), and achieves state-of-the-art results on multiple spatial benchmarks (top-right). We also uncover clear inter-level task dependencies in spatial supervised fine-tuning (bottom-right), offering guidance for designing future 3D spatially intelligent VLMs.}
\label{fig:teaser}
\vspace{-15pt}
\end{strip}

\input{sec/0_abstract}    
\input{sec/1_intro}
\input{sec/2_related}
\input{sec/3_method}

\input{sec/4_experiments}

\section{Conclusion}
We proposed a hierarchical framework organizing 3D spatial intelligence into a four-level taxonomy, progressively from basic geometric perception to complex abstract reasoning. Based on this framework, we build an automated pipeline and a large-scale dataset of diverse spatial VQA pairs from in-the-wild images and 3D-annotated data, enabling VLMs to learn comprehensive spatial understanding and reasoning via supervised fine-tuning. We also introduce an RGB-D VLM that incorporates metric-scale 3D point maps to enhance spatial understanding. Extensive experiments have demonstrated our state-of-the-art performance across diverse spatial benchmarks. Furthermore, our analysis highlighted clear inter-level dependencies among tasks, providing valuable insights for future training strategies to advance VLMs’ spatial intelligence.

{
    \small
    \bibliographystyle{ieeenat_fullname}
    \bibliography{main}
}


\input{sec/X_suppl}

\end{document}

%% file: sec/0_abstract.tex
\begin{abstract}

Achieving human-like spatial intelligence for vision-language models (VLMs) requires inferring 3D structures from 2D observations, recognizing object properties and relations in 3D space, and performing high-level spatial reasoning. In this paper, we propose a principled hierarchical framework that decomposes the learning of 3D spatial understanding in VLMs into \textbf{four progressively complex levels}, from geometric perception to abstract spatial reasoning. Guided by this framework, we construct an automated pipeline that processes approximately \textbf{5M images with over 45M objects} to generate 3D spatial VQA pairs across diverse tasks and scenes for VLM supervised fine-tuning. We also develop an \textbf{RGB-D VLM incorporating metric-scale point maps} as auxiliary inputs to further enhance spatial understanding. Extensive experiments demonstrate that our approach achieves state-of-the-art performance on multiple spatial understanding and reasoning benchmarks, surpassing specialized spatial models and large proprietary systems such as Gemini-2.5-pro and GPT-5. Moreover, our analysis reveals clear \textbf{dependencies among hierarchical task levels}, offering new insights into how multi-level task design facilitates the emergence of 3D spatial intelligence.

\end{abstract}

%% file: sec/1_intro.tex
\section{Introduction}
\label{sec:intro}

Vision-language models (VLMs) have demonstrated remarkable progress on a wide range of 2D vision-language tasks, including visual question answering (VQA)~\cite{liu2023visual,lu2024deepseek,bai2023qwen}, image captioning~\cite{yu2022coca,hu2022scaling}, visual grounding~\cite{steiner2024paligemma,xiao2024florence}, and action recognition~\cite{li2023videochat,wang2024internvideo2}. Extending these models from 2D perception to 3D spatial understanding, however, remains highly non-trivial, as it requires a holistic comprehension of 3D structures, object relations, and spatial layouts. Recent studies have attempted to equip VLMs with 3D reasoning abilities by introducing spatially oriented VQA tasks for supervised fine-tuning (SFT)~\cite{cheng2024spatialrgpt,chen2024spatialvlm,ma2025spatialllm,daxberger2025mm,cai2025spatialbot,liu2025ssr} or reinforcement fine-tuning (RFT)~\cite{ma2025spatialreasoner,zhou2025roborefer,shen2025fine}. Despite these advances, two major challenges persist.

First, a \textit{unified and systematic} task design that supports holistic 3D spatial intelligence across multiple cognitive levels is still lacking. It remains unclear how to define a task hierarchy that comprehensively captures the diverse reasoning skills required and reveals their underlying relationships. Second, \textit{large-scale, diverse, and 3D-grounded} data are difficult to obtain. Existing datasets with ground-truth 3D annotations are often limited to indoor scenes~\cite{dai2017scannet,zhu2024llava,zhang2025flatland,lazarow2025cubify}, while large-scale web data~\cite{kuznetsova2020open,peng2023kosmos} lack explicit 3D supervision, making them insufficient for robust spatial training. Some prior works~\cite{cheng2024spatialrgpt,deng2025internspatial,zhou2025roborefer} have explored constructing spatial reasoning datasets from in-the-wild images, yet their task-level coverage is still limited.

We revisit the design of spatial-related tasks for SFT of VLMs, and identify three essential aspects of 3D spatial understanding: recognizing object locations and properties in 3D space, understanding spatial relationships between objects, and developing high-level spatial imagination and reasoning. In this work, we aim to cultivate these diverse abilities that collectively define spatial intelligence, and systematically investigate their interdependencies through experiments.
To this end, we conceptualize 3D spatial intelligence as a four-level cognitive hierarchy, reflecting the progression from low-level perception to high-level reasoning. At the lowest tier (\emph{Level 0}), the model focuses on inferring 3D geometry directly from visual inputs, corresponding to fundamental geometric perception (\eg, monocular depth estimation). At \emph{Level 1}, the focus shifts to understanding intrinsic object-level 3D properties such as position, size, and orientation. Building upon these foundations, \emph{Level 2} emphasizes reasoning about spatial relationships among multiple objects to construct coherent 3D scene representations. Finally, \emph{Level 3} involves abstract spatial reasoning that integrates preceding abilities to support multi-step reasoning, mental simulation, and complex spatial problem-solving. Many existing studies can be positioned within one or more of these four levels according to their task design. 

Guided by this principle, we construct a diverse suite of spatial VQA tasks on generic images that span four cognitive levels of 3D spatial intelligence. To support such training, we develop an automated data generation pipeline that synthesizes hierarchical spatial VQA tasks from large-scale web data~\cite{peng2023kosmos,kakaobrain2022coyo-700m}, complemented by existing 3D-annotated datasets~\cite{lazarow2025cubify}. Ultimately, this pipeline processes roughly five million real-world images and over 45 million objects to generate a massive-scale corpus of more than two billion QA pairs, providing broad environmental and hierarchical coverage for supervised fine-tuning of VLMs.

Building on this foundation, we further design an RGB-D VLM, that integrates metric-scale 3D point maps as auxiliary input—obtained either from off-the-shelf monocular geometry estimators or ground-truth depth when available—to enhance spatial reasoning. Together, these contributions establish a comprehensive framework for developing and analyzing 3D spatial intelligence in VLMs. Our approach achieves state-of-the-art performance across multiple qualitative and quantitative spatial reasoning benchmarks, including CV-Bench~\cite{tong2024cambrian}, EmbSpatial~\cite{du2024embspatial}, 3DSRBench~\cite{ma20253dsrbench}, RoboSpatial~\cite{song2025robospatial}, SpatialRGPT~\cite{cheng2024spatialrgpt}, and QSpatial~\cite{liao2024reasoning}, outperforming both existing spatial specialist models and proprietary large models such as Gemini-2.5-pro~\cite{comanici2025gemini} and GPT-5~\cite{openai2025gpt5}, despite using only 3 billion parameters. Furthermore, our experiments reveal a \emph{clear hierarchical dependency} across task levels—incorporating lower-level tasks consistently enhances higher-level reasoning—offering new insights into the design of future training strategies for 3D spatially intelligent VLMs.

Our \textbf{main contributions} can be summarized as follows: 
\begin{itemize}
    \item We formulate 3D spatial intelligence into four carefully designed hierarchical levels and develop an automated data generation pipeline to construct a large-scale dataset with broader coverage of spatial understanding for VLMs beyond prior work.
    \item We develop a pointmap-augmented RGB-D VLM and finetune it on this dataset, achieving comprehensive and generalizable 3D spatial understanding with state-of-the-art results across various benchmarks.
    \item We uncover clear correlations among different spatial levels, offering principled insights for future training strategies to advance VLMs’ 3D spatial intelligence.
\end{itemize}

%% file: sec/2_related.tex
\section{Related Works}

\paragraph{Spatial understanding and reasoning with VLMs.} Spatial understanding and reasoning require perceiving object attributes, inferring spatial relationships and performing high-level reasioning about 3D scenes. While recent VLMs demonstrate impressive capability in 2D multimodal tasks~\cite{liu2023visual,li2023blip,alayrac2022flamingo,xiao2024florence,bai2025qwen2}, their ability to infer and reason about 3D spatial structures from 2D inputs remains limited~\cite{wang20233d,kamath2023s,shiri2024empirical,fu2024blink}. To address this, existing approaches~\cite{cho2024language,chen2024spatialvlm,cheng2024spatialrgpt,ma2025spatialllm,tong2024cambrian,zhu2024llava,zhou2025roborefer,ma2025spatialreasoner,deng2025internspatial, li2025spatialladderprogressivetrainingspatial, wang2025n3d, zhou2025robotracer, cai2025scaling, yang2025visual, gao2026holi, yang2025cambrians} integrate spatially-relevant VQA tasks into training, covering both qualitative and quantitative types. \textit{Qualitative} tasks mainly focus on relational reasoning, \eg, comparing object distance, orientations, or relative positions~\cite{kamath2023s,tong2024cambrian,wang2025spatialclip,ma2025spatialllm,song2025robospatial}, but language-based supervision alone often fails to capture fine-grained 3D structure. \textit{Quantitative} tasks, as in SpatialVLM~\cite{chen2024spatialvlm} and SpatialRGPT~\cite{cheng2024spatialrgpt}, involve metric-scale predictions such as distances, coordinates, or orientations~\cite{chen2024spatialvlm,cheng2024spatialrgpt,daxberger2025mm,zhou2025roborefer}, providing more precise spatial grounding. Beyond static observations, an emerging line of research extends spatial understanding to video and multi-view contexts~\cite{yang2025thinking, yin2025spatial, yang2025cambrians, damo2026rynnbrain, gao2026holi, yang2025visual, yang2025visual, wu2025spatial, cheng2025sr3d, xu2025multi}, challenging models to establish cross-view correspondence and comprehend dynamic spatio-temporal structures. More recent efforts combine higher-level reasoning tasks with reinforcement fine-tuning~\cite{rafailov2023direct,shao2024deepseekmath,zhou2025roborefer,ma2025spatialreasoner,shen2025fine,team2025robobrain,liao2025improved, liu2025spatial, yang2025visual, zhou2025robotracer, wang2025n3d, li2025spatialladderprogressivetrainingspatial}, enabling models to plan or simulate spatial relations. Despite these advances, a systematic hierarchical framework for designing spatial tasks remains underexplored. Some studies discuss spatial reasoning from a hierarchical perspective~\cite{wang2025spatial457,zheng2025multimodal} but focus mainly on evaluation. Others~\cite{cai2025spatialbot,zhang2025flatland} propose hierarchical task groupings, yet their level coverage and inter-level dependencies remain incomplete. Our work fills this gap by proposing a unified four-level hierarchy of spatial understanding and explicitly analyzing the dependencies among different task levels.

\paravspace
\paragraph{VLMs with auxiliary 3D information.} While a large body of work~\cite{hong20233dllm,zhu2024scanreason,chen2024ll3da,fu2024sceneLLM,huang2024leo,mao2025spatiallm} address holistic 3D scene understanding from full point-clouds scenes~\cite{dai2017scannet,zheng2020structured3d}, here we focus on methods that infer 3D spatial knowledge from monocular images. Extracting fine-grained 3D information from a single image is challenging, and several methods address this by incorporating 3D cues to enhance VLMs. Some approaches~\cite{cheng2024spatialrgpt,daxberger2025mm,liu2025ssr,chen2025sd,zhou2025roborefer} use \textit{relative depth} maps as auxiliary inputs, whereas~\cite{cai2025spatialbot} employs dedicated depth encoding to preserve metric-scale information. Others~\cite{zhu2024llava,zheng2025video,wu2025spatial,fan2025vlm,chen2025reasoning, cheng2025sr3d} leverage visual foundation models ~\cite{wang2025continuous, wang2025vggt} or design specialized additional encoders to accommodate the supplementary 3D spatial information derived from video or \textit{multi-view} inputs. We introduce a monocular RGB-D VLM that incorporates a metric-scale 3D point map--obtained from off-the-shelf monocular geometry estimators or sensor measurement -- as auxiliary input. It enhances spatial reasoning accuracy and achieves superior performance compared with models relying solely on relative depth information.

\begin{figure*}[t]
    \centering
    \includegraphics[width=\textwidth]{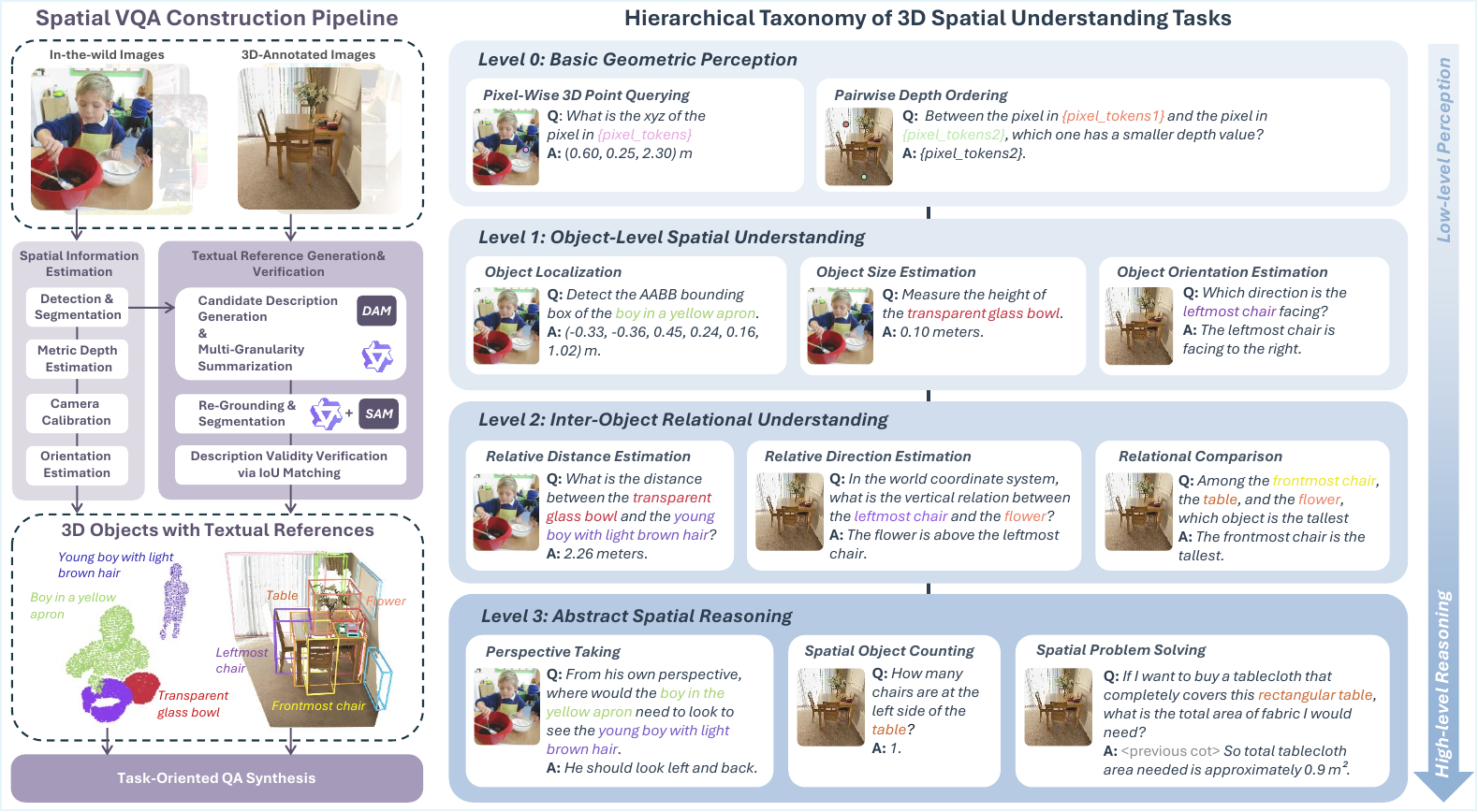}
    \caption{Overview of our approach. \textbf{Left:} Data construction pipeline which generates spatial-related VQA pairs from either in-the-wild images or existing data with 3D annotations. \textbf{Right:} Hierarchical spatial understanding task taxonomy with representative QA pairs.}
    \label{fig:overview}
    \vspace{-4pt}
\end{figure*}

%% file: sec/3_method.tex
\section{Method}

We aim to develop a VLM that acquires comprehensive 3D spatial understanding and reasoning capabilities from monocular visual input. To this end, we first introduce a principled hierarchical framework that progressively decomposes spatial-relevant tasks into hierarchical levels (Sec.~\ref{sec:task}). Guided by this principle, we constructed an automated data pipeline that generates diverse 3D spatial QA pairs from both in-the-wild images and datasets with 3D annotations (Sec.~\ref{sec:data_pipe}). Finally, we design a VLM that integrates a metric-scale point map and train it on our large-scale dataset to achieve holistic 3D spatial intelligence (Sec.~\ref{sec:model_architecture}). An overview is shown in Fig.~\ref{fig:overview}.

\subsection{Hierarchical 3D Spatial Understanding}\label{sec:task}

To capture the progressive nature of spatial understanding, we organize spatial-related VQA tasks into four hierarchical levels, each building upon the previous one and reflecting a transition from low-level geometric perception to high-level spatial reasoning. The following paragraphs describe these four levels in detail and Fig.~\ref{fig:overview} presents examples.

\paravspace
\paragraph{Level 0: Basic geometric perception.} 
Perceiving depth and inferring 3D structure from 2D visual input are fundamental aspects of human spatial intelligence~\cite{welchman2016human} and have long been central problems in computer vision~\cite{agarwal2011building,schonberger2016pixelwise,yang2024depth}. To endow VLMs with similar capabilities, we accordingly design two VQA tasks: 

\vspace{3pt}
\noindent{\textit{(i) Pixel-wise 3D point querying}}, which requires the model to output the metric-scale 3D coordinates of a given 2D image location in the camera coordinate system. 

\vspace{3pt}
\noindent{\textit{(ii) Pairwise depth ordering}}, aiming to determine the relative depth between two given image coordinates. 

\vspace{3pt}
These tasks involve basic geometric perception abilities without relying on specific semantic information.

\paravspace
\paragraph{Level 1: Object-level spatial understanding.} Building upon geometric perception, this level integrates semantic grounding with spatial localization. The model must not only perceive 3D geometry but also associate it with object identity and meaning. It learns to link linguistic or visual references to objects and infer their 3D spatial attributes, such as position, size, and orientation. These capabilities bridge \textit{perception and semantics}, enabling reasoning about discrete entities in the 3D world:  

\vspace{3pt}
\noindent{\textit{(i) Object localization}}, predicting the 3D position of an object in the camera coordinate system via its bounding box. 

\vspace{3pt}
\noindent{\textit{(ii) Object orientation estimation}}, which describes the object’s yaw direction via linguistic references (\eg, front/back, left/right, up/down). 

\vspace{3pt}
\noindent{\textit{(iii) Object size estimation}}, estimating the object’s physical dimensions, such as width and height.

\vspace{3pt}
For the objects involved in these tasks, the model receives either a linguistic description or a visual reference (a 2D bounding box), enabling flexible object grounding, and produces the corresponding quantitative metrics or qualitative descriptions (see Fig.~\ref{fig:overview} for example outputs).

\paravspace
\paragraph{Level 2: Inter-object relational understanding.} Given an understanding of individual objects and their 3D attributes, the next level focuses on relationships among multiple objects. Here, the model must integrate the object-level representations from Level 1 and reason jointly about their relative positions, orientations, and distances, forming a complete scene representation. We define three representative tasks:

\vspace{3pt}
\noindent{\textit{(i) Relative direction estimation}}, which evaluates relative placement between two objects in camera frame. The model is required to predict either a qualitative estimate (\eg, left/right, front/behind, below/above) or the precise 3D direction vector between the objects.

\vspace{3pt}
\noindent{\textit{(ii) Relative distance estimation}}, which quantitatively predicts the relative distance between objects from multiple perspectives, including Euclidean distance, as well as vertical, horizontal, and depth-wise components. 

\vspace{3pt}
\noindent{\textit{(iii) Relational comparison}}, which compares multiple objects ($\ge 2$) based on a shared attribute, including position, orientation, or size. For position- and size-related tasks, this involves selecting objects with extreme values (\eg, nearest/farthest, smallest/largest) or ordering objects according to the attribute. Orientation-related tasks focus on assessing directional consistency between objects (\eg, similar, orthogonal, or opposite).

\paravspace
\paragraph{Level 3: Abstract spatial reasoning.} On top of relational reasoning, this level targets high-order inference that goes beyond directly perceivable relations, where the model should generalize spatial knowledge, imagine alternative viewpoints, and reason towards implicit goals—paralleling human abstract reasoning~\cite{national2005learning,tversky2009embodied,piaget2013child}. We design three types of tasks for the model to acquire such ability:

\vspace{3pt}
\noindent \textit{(i) Perspective taking}, which requires the model to infer the relative directions and distances of objects from an imagined observer- or object-centric viewpoint. 

\vspace{3pt}
\noindent \textit{(ii) Spatial object counting}, which requires the model to identify and enumerate target objects that satisfy specific spatial relationship constraints relative to a given reference anchor.

\vspace{3pt}
\noindent \textit{(iii) Spatial problem solving}, which involves inferring spatial attributes from a high-level objective, translating the objective into quantifiable spatial properties, and performing multi-step reasoning and computation to determine a solution.

\vspace{6pt}
\noindent These tasks cover a broad spectrum of spatial understanding and reasoning abilities necessary for the model. To realize them at scale, we design an automated data pipeline that constructs the corresponding VQA pairs from in-the-wild images or images with 3D annotations, as described below.

\subsection{Spatial VQA Data Construction}\label{sec:data_pipe}

Our data pipeline consists of three main stages: spatial information extraction, textual reference generation, and task-oriented QA synthesis (as illustrated in Fig.~\ref{fig:overview}).

\paravspace

\paragraph{Spatial information estimation.} We estimate metric-scale 3D spatial information from 2D images by generating pixel-wise 3D point maps using MoGe-2~\cite{wang2025moge}. Our object localization pipeline adapts to available annotations. For unannotated data, following~\cite{cheng2024spatialrgpt,zhou2025roborefer}, we sequentially apply RAM~\cite{zhang2024recognize} for categorization, GroundingDINO~\cite{liu2024grounding} for 2D bounding boxes, and SAM~\cite{ravi2024sam,kirillov2023segment, carion2025sam} for masking; for data with existing 2D bounding boxes, we bypass the first two steps and directly prompt SAM using the ground-truth boxes. Combining these masks with the 3D point map yields object-level point clouds to derive 3D bounding boxes and sizes. Object orientations are estimated via OrientAnythingv2~\cite{wang2026orient}. Finally, we establish a gravity-aligned world coordinate system ($y$-axis parallel to gravity) using Perspective Fields~\cite{jin2023perspective} to compute relative spatial relationships. If ground-truth 3D annotations are available, this entire estimation pipeline is skipped.

\paravspace
\paragraph{Textual reference generation.} 
To generate textual references for individual objects in the scene, we leverage a combination of Describe Anything~\cite{lian2025describe}, Qwen2.5-VL~\cite{Qwen2.5-VL}, and Qwen3-VL~\cite{Qwen3-VL} to obtain comprehensive object descriptions.

Since the generated textual descriptions may contain errors or ambiguities (\eg, multiple objects in the image matching the same description), we introduce an additional verification process to enhance the accuracy of object referencing. Specifically, we prompt VLM to ground each object using its generated description and evaluate the correspondence between the predicted bounding box and the original one (obtained from the previous stage). Results with an IoU below a certain threshold are considered invalid, and the corresponding textual descriptions are discarded. If none of an object’s textual references pass the verification stage, we instead describe it using its class label combined with a visual cue, \ie, the textual reference \emph{``[object\_class] (highlighted by [color] box)"} along with its corresponding 2D bounding box on the image. Further details can be found in the Appendix.

\paravspace
\paragraph{Task-oriented QA synthesis.}
Based on the available spatial information and textual references, we generate QA instances following the hierarchical task taxonomy described in Sec.~\ref{sec:task}. 
To enhance diversity and provide complementary learning signals, each task type is generated in three template formats whenever applicable: 

\vspace{4pt}
\noindent{\textit{(i) Free-form question answering}}, enabling the model to respond to questions in an open-ended manner. 

\vspace{4pt}
\noindent{\textit{(ii) Multiple-choice questions (MCQs)}}, where the model selects the correct answer from several options.

\vspace{4pt}
\noindent{\textit{(iii) True/False}}, which assesses a statement’s correctness.

\vspace{4pt}
As an exception, for the Level-3 \emph{spatial problem solving} tasks, we do not use templates. Instead, we provide GPT with the original image, corresponding spatial information, and textual references as prompts, instructing it to formulate questions that require multi-step reasoning over object properties and spatial relations, thereby generating more complex reasoning examples.

\begin{figure}[t]
    \centering
    \includegraphics[width=\linewidth]{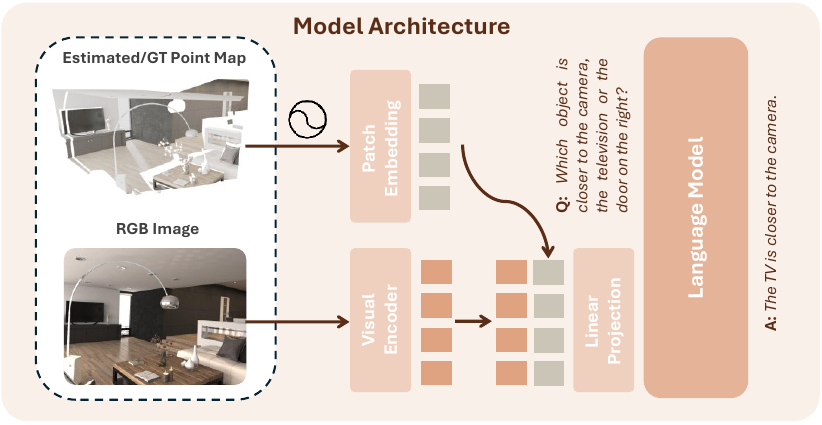}
    \caption{Model architecture of our VLM, which integrates metric-scale 3D point map as auxiliary input.}
    \label{fig:model}
    \vspace{-4pt}
\end{figure}

\paravspace

\paragraph{Data sources.} We construct our training dataset from three primary sources: KosMos-2~\cite{peng2023kosmos}, Objects365~\cite{shao2019objects365}, and CA-1M~\cite{lazarow2025cubify}. KosMos-2, a subset of COYO-700M~\cite{kakaobrain2022coyo-700m}, contains 15M in-the-wild images, from which we filtered 3.8M for VQA data generation. Objects365 provides 1M in-the-wild images with ground-truth object bounding boxes. CA-1M offers 2M indoor video frames with dense point clouds and 3D boxes, from which we sampled 200K images. In total, we curated a large-scale spatial VQA dataset comprising \textbf{5M images}, \textbf{45M objects}, and \textbf{2B QA pairs} spanning diverse formats. This dataset is then utilized for supervised fine-tuning of our VLM, as detailed below.

\begin{table*}[t]
  \caption{Accuracy (\%) on \textbf{quantitative} VQA benchmarks for level-1 and level-2 spatial understanding. $*$ denotes using GT point map.}
  \vspace{-2pt}
  \label{tab:quant}
  \centering
  \footnotesize
  \setlength{\tabcolsep}{2.5pt}
  \scalebox{1}{
  \begin{tabular}{@{}p{4cm}cccccccccc@{}}
    \toprule
    \textbf{Model} & \textbf{Input} &
    \multicolumn{6}{c}{\textbf{SpatialRGPT-Quantitative} $\uparrow$}  &
    \multicolumn{3}{c}{\textbf{QSpatial-Bench} $\uparrow$} \\
     \cmidrule(lr){3-8} \cmidrule(lr){9-11}
     & &
    \raisebox{1ex}[0pt]{Width} &
    \raisebox{1ex}[0pt]{Height} &
    \shortstack{Direct\\Distance} &
    \shortstack{Horizontal\\Distance} &
    \shortstack{Vertical\\Distance} &
    \raisebox{1ex}[0pt]{\textbf{Avg.}} &
    \raisebox{1ex}[0pt]{Plus} &
    \raisebox{1ex}[0pt]{ScanNet} & 
    \raisebox{1ex}[0pt]{\textbf{Avg.}} \vspace{-8pt}\\
     & &  
     \multicolumn{2}{c}{\scalebox{0.8}{$\underbrace{\rule{2.0cm}{0pt}}_{\textit{\footnotesize{(Level 1)}}}$}} &
     \multicolumn{3}{c}{\scalebox{0.8}{$\underbrace{\rule{4.4cm}{0pt}}_{\textit{\footnotesize{(Level 2)}}}$}} &
     &
     \multicolumn{2}{c}{\scalebox{0.8}{$\underbrace{\rule{2.0cm}{0pt}}_{\textit{\footnotesize{(Level 1\&2)}}}$}} \\
   
    \midrule
    \multicolumn{10}{l}{\textit{Proprietary Models}} \\
    GPT-4o & RGB & 51.10 & 68.40 & 29.70 & 25.40 & 33.00 & 41.50 & 61.06  & 69.41 & 66.30 \\
    GPT-5 & RGB & 59.09 & 67.66 & 29.72 & 25.21 & 13.72 & 40.47 & 63.37 & 73.53
    & 68.45 \\
    Gemini-2.5-Pro & RGB & 37.20 & 43.51 & 16.78 & 17.39  & 15.53 & 26.57 & 40.59 & 55.46 & 49.92 \\
    \midrule
    \multicolumn{10}{l}{\textit{Open-Source Vision-Language Models}} \\
    PaliGemma2-3B & RGB & 20.3 & 27.07 & 20.27 & 15.57 & 22.64 & 21.18 & 36.63 & 30.06 & 32.84 \\
    Qwen3VL-8B & RGB & 27.06 & 40.60 & 31.10 & 31.96 & 18.86 & 30.37 & 35.64 & 55.88 & 48.34 \\
    InternVL3.5-8B & RGB & 27.81 & 41.35 & 31.08 & 31.96  & 17.92 & 30.53 & 59.41 & 57.06 & 57.94\\
    \midrule
    \multicolumn{10}{l}{\textit{Spatial Specialist Models}} \\
    RoboRefer-8B-SFT & RGB-D & 17.29 & 60.15 & 49.32 & 37.70 & 28.30 & 39.25 & 20.79 & 31.07 & 27.24 \\
    SpatialRGPT-8B & RGB-D & 48.90 & 61.70 & 45.90  & 68.00 & 56.60 & 56.22 & - & - & - \\
    MM-Spatial-3B & RGB-D & 55.60 & 83.50 & 59.50 & \underline{82.00} & 63.20 & 68.70 & - & - & -\\
    \midrule
    \multicolumn{10}{l}{\textit{Our Model Variants}} \\
    HiSpatial-3B-RGB & RGB & \textbf{69.92} & \textbf{84.96} & \underline{66.89} & 71.34 & \underline{68.87} & \underline{72.43} & \underline{76.24} & \underline{75.88} & \underline{76.01} \\
    HiSpatial-3B & RGB-XYZ & \underline{69.17} &\textbf{84.96} & \textbf{79.73} & \textbf{86.89} & \textbf{75.47} & \textbf{79.28} & \textbf{88.12} & \textbf{84.17} & \textbf{85.16} \\
    \textcolor{gray}{HiSpatial-3B$^*$} & \textcolor{gray}{RGB-XYZ (GT)} &
    \textcolor{gray}{\textbf{70.68}} &
    \textcolor{gray}{\underline{84.21}} &
    \textcolor{gray}{\textbf{83.11}} &
    \textcolor{gray}{\textbf{90.16}} &
    \textcolor{gray}{\textbf{79.25}} &
    \textcolor{gray}{\textbf{81.46}} & 
    - &
    \textcolor{gray}{\underline{84.12}} &
    -\\
    \bottomrule
  \end{tabular}
  }
  \vspace{-2pt}
\end{table*}

\subsection{Spatial VLM Finetuning}
\label{sec:model_architecture}

\paragraph{Model architecture.} Our VLM architecture is shown in Fig.~\ref{fig:model}. We build upon PaliGemma-2~\cite{steiner2024paligemma}, which integrates a SigLIP~\cite{zhai2023sigmoid} vision encoder with a Gemma-2~\cite{team2024gemma} language model. The original VLM processes only RGB images; we augment it with depth information to enhance spatial understanding. Unlike prior methods~\cite{zhou2025roborefer,cheng2024spatialrgpt,daxberger2025mm,liu2025ssr} that use relative depth, we employ a metric-scale 3D point map. This provides richer 3D information, leading to improved spatial reasoning, as shown in our experiments.

Specifically, the 3D point map $\mathbf{X} \in \mathbb{R}^{H \times W \times 4}$ stores each 2D pixel’s 3D coordinates in the camera frame in its first three channels, and the fourth channel is a binary mask indicating whether the point is valid. The point map undergoes sinusoidal positional encoding and a learnable patchify layer (Conv2D), resulting in a feature map with the same spatial dimensions as the RGB image features from the visual encoder. These two feature maps are concatenated along the feature dimension and passed through a linear projector to produce fused tokens, which serve as a replacement for the original visual input tokens of the language model. The textual information is concatenated after the fused visual tokens, and the model autoregressively generates text tokens to produce the corresponding answer.
By this design, the VLM can exploit metric 3D cues meanwhile maintain compatibility with its pretrained visual pathway. 

By default, we use MoGe-2 to estimate the 3D point map from the input RGB image. When ground-truth point maps (\eg, obtained from depth sensors) are available, they can be used instead to further improve performance.

\paravspace
\paragraph{Training objective.} Our training follows the standard SFT procedure of VLM~\cite{steiner2024paligemma}, minimizing the cross-entropy loss for each output token $\mathbf{y}_t$ given previous tokens $\mathbf{y}_{<t}$ and visual input (\ie, RGB image $\mathbf{I}$ and point map $\mathbf{X}$):
\begin{equation}
\mathcal{L} = - \sum_{t=1}^{T} \log P_\theta(\mathbf{y}_t \mid \mathbf{y}_{<t}, \mathbf{I}, \mathbf{X}). \label{eq:loss}
\end{equation}
For unified training across tasks via Eq.~\eqref{eq:loss}, all QA pairs from our data pipeline, both quantitative and qualitative, are converted into text descriptions. During training, the visual encoder is frozen, and the point map patchify layer, fused-token projector, and LLM are jointly fine-tuned end-to-end.

%% file: sec/4_experiments.tex
\begin{table*}[t]
  \caption{Accuracy (\%) on \textbf{qualitative} VQA benchmarks evaluating spatial understanding and reasoning across levels 1–3.}
  \vspace{-6pt}
  \label{tab:qualitative_results}
  \centering
  \footnotesize
  \setlength{\tabcolsep}{2pt}
  \scalebox{1}{
  \begin{tabular}{@{}p{4cm}ccccccc@{}}
    \toprule
    \textbf{Method}  & \textbf{Input} &
    \textbf{EmbSpatial} $\uparrow$ & \textbf{RoboSpatial} $\uparrow$ & \textbf{CV-Bench-3D} $\uparrow$ & \textbf{CV-Bench-2D Relation} $\uparrow$ & \textbf{3DSRBench} $\uparrow$ \vspace{-7pt}\\
     & &  \multicolumn{4}{c}{\scalebox{0.8}{$\underbrace{\rule{10.8cm}{0pt}}_{\textit{\footnotesize{(Level 2)}}}$}} & \multicolumn{1}{c}{\scalebox{0.8}{$\underbrace{\rule{2.0cm}{0pt}}_{\textit{\footnotesize{(Level 1-3)}}}$}} \\
    \midrule
    \multicolumn{7}{l}{\textit{Proprietary Models}} \\
    GPT-4o & RGB & 63.38 & 77.20 & 84.90 & 84.62 & 44.20 \\
    Gemini-2.5-Pro & RGB & 76.67 & 77.24 & 90.80 & 93.54 & 48.47 \\
    Claude-3.7-Sonnet & RGB & 33.33 & 60.73 & 85.00 & 74.15 & 48.20 \\
    \midrule
    \multicolumn{7}{l}{\textit{Open-Source Vision-Language Models}} \\
    PaliGemma2-3B & RGB & 28.32 & 71.54 & 42.17 & 72.00 & ~~8.15 \\
    InternVL3.5-8B & RGB & 69.81 & 78.86 & 85.09 & 88.00 & 51.58 \\
    InternVL3.5-14B & RGB & 71.62 & 78.86 & 86.75 & 88.15 & 57.32 \\
    Qwen-3-VL-8B & RGB & 78.50 & 82.11 & 90.66 & 92.92 & 52.80 \\
    Qwen-3-VL-30B-A3B & RGB & 76.40 & 83.73 & 92.00 & 95.38 & 55.70 \\
    \midrule
    \multicolumn{7}{l}{\textit{Spatial Specialist Models}} \\
    SpatialBot-3B & RGB-D & 50.66 & 72.36 & 69.08 & 69.38 & 54.67 \\
    SpaceLLaVA-13B & RGB & 49.40 & 61.00 & 68.50 & 63.69 & 46.55 \\
    SpatialRGPT-8B & RGB-D & 59.62 & 66.67 & 89.15 & 91.00 & 48.40 \\
    RoboRefer-8B-SFT & RGB-D & 72.53 & \underline{84.55} & \underline{95.92} & \textbf{96.90} & 51.24 \\
    \midrule
    \multicolumn{7}{l}{\textit{Our Model Variants}} \\
    HiSpatial-3B-RGB  & RGB & \underline{79.78} & 83.74 & 95.58 & 95.08 & \textbf{64.34} \\
    HiSpatial-3B & RGB-XYZ & \textbf{80.71} & \textbf{86.18} & \textbf{97.58} & \underline{95.69} & \underline{63.81} \\
    \bottomrule
  \end{tabular}
  } \label{tab:quali}
  \vspace{-4pt}
\end{table*}

\section{Experiments}\label{sec:experiments}

\paragraph{Implementation details.} We initialize our VLM from PaliGemma2-3B-Mix-448 with $448^2$ input resolution. During SFT, we combine our spatial VQA data with the general VQA data from LLaVA-Next~\cite{liu2024llavanext} to preserve general-purpose ability, using a sampling ratio of 1:7 (general : spatial). We train the model for up to 70K iterations with a batch size of 256, covering the dataset images for roughly 3 epochs though not all QA pairs are utilized. 
The AdamW~\cite{loshchilov2017decoupled} optimizer is used ana learning rate is set to $2\times10^{-5}$. More details are in the supplementary material.

\subsection{Evaluation Benchmarks}\label{sec:benchmarks}

\paragraph{Public spatial reasoning benchmarks.} We evaluate our model on public benchmarks across different task levels: SpatialRGPT~\cite{cheng2024spatialrgpt} and QSpatial~\cite{liao2024reasoning} (level 1–2); CV-Bench~\cite{tong2024cambrian}, EmbSpatial~\cite{du2024embspatial}, and RoboSpatial~\cite{song2025robospatial} (level 2); and 3DSRBench~\cite{ma20253dsrbench} (level 1–3). SpatialRGPT and QSpatial use quantitative rules: predictions within $0.75$–$1.25\times$ of the ground truth (SpatialRGPT) or within $0.5$–$2\times$ (QSpatial) are considered correct, and accuracy is reported.
The remaining benchmarks use multiple-choice or judgment-based qualitative evaluation and report accuracy as well.

\paravspace
\paragraph{Custom benchmark.} Existing public spatial reasoning benchmarks cover only part of our model’s capabilities. To provide a more comprehensive evaluation, we design a custom benchmark spanning levels 1–3 using 3D-annotated Omni3D~\cite{brazil2023omni3d} data and the CA-1M test set. For the level-1 task, we estimate \emph{object-to-camera distance} on 307 questions, computing accuracy as in SpatialRGPT. For level-2, we evaluate \emph{relative direction} between two objects on 302 questions, counting predictions (unit vectors in the camera frame) as correct if within $30^\circ$ of the ground truth.

\begin{table}[t]
  \caption{Accuracy (\%) on our custom spatial VQA benchmark.}
  \vspace{-6pt}
  \label{tab:additional_results_transposed}
  \centering
  \footnotesize
  \setlength{\tabcolsep}{4pt}
  \scalebox{0.9}{
  \begin{tabular}{lcccc}
    \toprule
    \textbf{\shortstack{Model}} & 
    \!\textbf{\shortstack{\!Object-to-Camera\!\\Distance (L1) }}\!\!\!$\uparrow$\! & 
    \!\textbf{\shortstack{Object\\Direction (L2)}} $\uparrow$\! & 
    \!\textbf{ \shortstack{Spatial Problem \\ Solving (L3)}}\!\!\!$\uparrow$\! \\
    \midrule
    GPT-5              & 47.19 & \underline{59.27} & \underline{33.33} \\
    Qwen3VL-8B         & 12.70 & 22.52 & 25.64 \\
    RoboRefer-8B-SFT   & \underline{58.63} & N/A & 26.92 \\
    HiSpatial-3B (ours)               & \textbf{92.18} & \textbf{67.21} & \textbf{47.44} \\
    \bottomrule
  \end{tabular}
  }\label{tab:self}
  \vspace{-2pt}
\end{table}

For level 3, we introduce a \emph{spatial problem-solving} task with 78 diverse questions. It evaluates the model’s ability to connect abstract, requirement-based questions with a scene’s spatial properties and perform multi-step reasoning and computation (as in Fig.~\ref{fig:overview}). 
For evaluation, we use GPT-4.1 to extract keywords and assess answer correctness, following~\cite{cheng2024spatialrgpt}. For judgment-based questions, accuracy is computed directly, while for quantitative questions, predictions within $25\%$ of the ground truth are considered correct. See the supplementary material for details.

\paravspace
\paragraph{General VQA benchmarks.} 
We further evaluate the model’s performance on general real-world visual understanding after spatial-task fine-tuning using several general VQA benchmarks, including MMBench~\cite{liu2024mmbench}, POPE~\cite{li2023evaluating}, SEED~\cite{li2023seed}, and RealWorldQA~\cite{xia2024realworldqa}.

\begin{table}[t]
  \caption{Accuracy (\%) on general VQA benchmarks compared to our base model PaliGemma2.}
  \vspace{-6pt}
  \label{tab:general_benchmarks}
  \centering
  \footnotesize
  \setlength{\tabcolsep}{4pt}
  \scalebox{1}{
  \begin{tabular}{@{}lcc@{}}
    \toprule
    \textbf{Benchmark} & \textbf{PaliGemma2-3B} & \textbf{Ours} \\
    \midrule
    MMBench $\uparrow$ & 49.86 & \textbf{69.67} \\
    POPE $\uparrow$  & 87.00 & 
    \textbf{87.97} \\
    SEED $\uparrow$ & 48.32 & \textbf{63.51} \\
    RealWorldQA $\uparrow$ & 47.76 & \textbf{58.95} \\
    \bottomrule
  \end{tabular}
  }
  \vspace{-2pt}
\end{table}

\begin{table*}[t]
\centering
\footnotesize
\caption{Inter-level task dependency analysis. Removing lower-level tasks in training reduces higher-level performance; see text for details.}
\vspace{-5pt}
\label{tab:ablation_task}
\setlength{\tabcolsep}{4pt}

\begin{tabular}{cccc|ccccc|ccc}
\toprule
\textbf{L0} & \textbf{L1} & \textbf{L2} & \textbf{L3}
& \multicolumn{5}{c}{\textbf{Level 2 Tasks}} 
& \multicolumn{3}{c}{\textbf{Level 3 Tasks}} \\
\cmidrule(lr){5-9}\cmidrule(lr){10-12}
& & &
& CV-Bench $\uparrow$ & RoboSpatial $\uparrow$ & 3DSR-L2 $\uparrow$ & EmbSpatial $\uparrow$ & \emph{Avg.} $\uparrow$
& 3DSR-L3 $\uparrow$ & \!\!Problem Solving $\uparrow$\!\! & \emph{Avg.} $\uparrow$ \\
\midrule

\checkmark & \checkmark & \checkmark & \checkmark
& 96.64 & 86.18 & 61.32 & 80.71 & 81.21
& 65.14 & 47.44 & 56.29 \\
\cmidrule(lr){5-12}

& & \checkmark & \checkmark
& \!96.55 \textcolor{blue}{\scriptsize{\ \ (-0.09)}}\! & \!82.93 \textcolor{blue}{\scriptsize{(-3.25)}}\! & \!60.03 \textcolor{blue}{\scriptsize{\ \ (-1.29)}}\! & \!79.25 \textcolor{blue}{\scriptsize{\ \ (-1.46)}}\! & \!79.69 \textcolor{blue}{\scriptsize{\ \ (-1.52)}}\!
& \!51.43 \textcolor{blue}{\scriptsize{(-13.71)}}\! & \!44.87 \textcolor{blue}{\scriptsize{(-2.57)}}\! & \!48.15 \textcolor{blue}{\ \ \scriptsize{(-8.14)}}\! \\

\checkmark & & & \checkmark
& \!79.00 \textcolor{gray}{\scriptsize{(-17.64)}}\! & \!77.24 \textcolor{gray}{\scriptsize{(-8.94)}}\! & \!37.36 \textcolor{gray}{\scriptsize{(-23.96)}}\! & \!37.53 \textcolor{gray}{\scriptsize{(-43.18)}}\! & \!56.21 \textcolor{gray}{\scriptsize{(-25.00)}}\! & \!43.81 \textcolor{blue}{\scriptsize{(-21.33)}}\! & \!39.74 \textcolor{blue}{\scriptsize{(-7.70)}}\! & \!41.78 \textcolor{blue}{\scriptsize{(-14.51)}}\!\\

\bottomrule
\end{tabular}
\vspace{-4pt}
\end{table*}

\begin{table}[t]
  \caption{Effect of auxiliary 3D input on model accuracy (\%).}
  \vspace{-5pt}
  \label{tab:ablation_input}
  \centering
  \footnotesize
  \setlength{\tabcolsep}{4pt}
  \scalebox{1}{
  \begin{tabular}{@{}lcccc@{}}
    \toprule
    \textbf{Input} & \textbf{Source} & \textbf{Qualitative $\uparrow$}  & \textbf{Quantitative $\uparrow$} \\ 
    \midrule
    RGB & - & 83.70 ~~~~~~~~~\,\, & 74.16 ~~~~~~~~~\,\, \\
    RGB+Relative Depth & MoGe2 & 84.29 \textcolor{blue}{\scriptsize{(+0.59)}} & 75.26 \textcolor{blue}{\scriptsize{(+0.90)}} \\

    RGB+XYZ (Ours) & MoGe2 & \textbf{84.79} \textcolor{blue}{\scriptsize{(+0.50)}} & \textbf{82.02} \textcolor{blue}{\scriptsize{(+6.76)}} \\
    \midrule
    \textcolor{gray}{RGB+XYZ (Ours)} & \textcolor{gray}{GT} & - &  \textcolor{gray}{82.79} {\textcolor{gray}{\scriptsize{(+0.77)}}} 
    \\
    \bottomrule
  \end{tabular}
  }
  \vspace{-4pt}
\end{table}

\subsection{Spatial Understanding Evaluation}

\paragraph{Baselines.} We compare our approach on the above benchmarks against spatially specialized models (SpatialBot~\cite{cai2025spatialbot}, SpaceLLaVA~\cite{chen2024spatialvlm}, SpatialRGPT~\cite{cheng2024spatialrgpt}, RoboRefer~\cite{zhou2025roborefer}, and MM-Spatial~\cite{daxberger2025mm}), general open-source models (PaliGemma2~\cite{steiner2024paligemma}, Qwen3-VL~\cite{qwen3vl2024} and InternVL3.5~\cite{wang2025internvl3}), and proprietary models (GPT-4o~\cite{achiam2023gpt}, GPT-5~\cite{openai2025gpt5}, Gemini-2.5-Pro~\cite{comanici2025gemini}, and Claude-3.7-Sonnet~\cite{anthropic2024claude37sonnet}).

\paravspace
\paragraph{Results.}
Figure~\ref{fig:teaser} demonstrates qualitative results of our method. Table~\ref{tab:quant} reports quantitative VQA results for levels 1 and 2, Table~\ref{tab:quali} shows qualitative VQA performance for levels 1–3, and Table~\ref{tab:self} summarizes results on our custom benchmark covering levels 1–3.

Our method substantially outperforms previous approaches on quantitative tasks across levels 1 and 2 (Table~\ref{tab:quant}). Even without point map input, our variant surpasses existing general VLMs, including GPT-5 and Gemini-2.5-Pro, and achieves comparable performance to the best spatial specialist models that use additional depth inputs. Incorporating an auxiliary point map (estimated via MoGe-2) further improves performance, and using ground-truth depth for the point map boosts results even more, highlighting strong potential for downstream tasks with depth sensors, such as embodied AI scenarios. Table~\ref{tab:quali} shows that our method achieves state-of-the-art performance on multiple qualitative benchmarks, demonstrating comprehensive spatial understanding and reasoning skills. Compared to our base model (PaliGemma2-3B), we achieve substantial gains in both quantitative and qualitative metrics, demonstrating the effectiveness of our spatial data.

Table~\ref{tab:self} highlights the broad coverage of our method in spatial understanding and reasoning. Our model shows substantial gains over others on quantitative tasks at levels 1 and 2 and outperforms existing approaches on level-3 problem-solving tasks requiring complex spatial reasoning.

\subsection{General VQA Evaluation}

In \cref{tab:general_benchmarks}, we compare our model with its base version to assess the impact of spatial supervised fine-tuning. The results show that our model retains its general abilities, even surpassing the original VLM. Notably, during fine-tuning, 88\% of the data comes from spatial tasks and 12\% from general VQA, indicating that enhancing spatial understanding does not compromise general VQA performance, consistent with prior observations~\cite{zhou2025roborefer,chen2024spatialvlm}.

\subsection{Ablation Studies and Analysis}\label{sec:ablation}

\paragraph{Inter-level task dependencies.} Understanding interactions between tasks at different levels is key for designing fine-tuning strategies to enhance spatial intelligence, yet this has been largely overlooked in previous approaches. To investigate this, we conduct ablation studies by selectively removing VQA data from specific levels and measuring the effect on the model’s performance at other levels. Two baselines are designed by removing tasks from levels 0 \& 1 and levels 1 \& 2, respectively; {the missing data is then backfilled with general VQA data to ensure that \emph{the number of the other levels' spatial reasoning samples seen by the model remains unchanged}}.

As shown in Table~\ref{tab:ablation_task}, removing tasks from levels 0 and 1 during training leads to a clear performance drop on level 2. Notably, level 2 tasks are more numerous than those in levels 0 and 1. Furthermore, they do not explicitly rely on chain-of-thought (CoT) reasoning from levels 0 and 1. Despite this, removing the lower-level tasks still hurts level 2 performance, suggesting that levels 0 and 1 help the VLM implicitly capture richer spatial information that benefits higher-level tasks.

Regarding level 3, removing tasks from levels 0 and 1 or from levels 1 and 2 both lead to a significant performance drop, with removing levels 1 and 2 causing a much larger decrease (-14.51\% \emph{vs.} -8.14\%) than removing levels 0 and~1. This is because level 3 tasks are more complex and rely more directly on the specific skills developed in levels 1 and 2. Additionally, since level 3 has less training data, the model depends more heavily on the knowledge transferred from these intermediate levels. Without this hierarchical support, the model's performance on high-level spatial reasoning is greatly compromised.

\paravspace
\paragraph{Influence of auxiliary 3D input.} We evaluate our default architecture with metric-scale point maps against an alternative that replaces the point map with relative depth maps, as in most prior work~\cite{zhou2025roborefer, liu2025ssr, cheng2024spatialrgpt, daxberger2025mm}.

Table~\ref{tab:ablation_input} reports average performance across spatial benchmarks. Incorporating metric-scale point maps enhances spatial understanding more effectively than relative depth. On quantitative tasks, incorporating ground-truth point maps yields additional performance improvements by providing the exact metric scale for accurate estimation. Consequently, our framework can flexibly exploit GT information in scenarios where depth data is available, further boosting spatial understanding.

%% file: sec/X_suppl.tex
\newpage

\section{More Implementation Details}
\subsection{Model Architecture}
In this section, we provide  detailed implementation information for our RGB-D VLM, including the parameter settings for the 3D input branch and the inference procedure for incorporating ground-truth point maps.

\paravspace
\paragraph{Auxiliary 3D input branch.} The input to our 3D branch is a 3-channel metric point map, where the point values lie within the range $[-250,250]$ (meters).
We apply a 64-dimension sinusoidal positional encoding to each coordinate, resulting in a total of 192 channels. After concatenating an extra 1-channel validity mask, the input becomes a 193-channel feature map, which is then passed through a Conv2D layer with a $14^2$ kernel and a stride of 14 to downsample it to the spatial resolution of the RGB feature map from the SigLIP encoder. The downsampled 3D feature map is concatenated with the RGB feature map along the channel dimension, resulting in a final feature map of 2304 channels. This combined feature map is then passed through a linear projection layer to match the embedding space of the language model for subsequent processing.

\paravspace
\paragraph{Inference with ground-truth point map.} While our RGB-D VLM uses the point map estimated by MoGe-2 as the default input, it can also utilize ground-truth point maps when they are available. However, the ground-truth depth maps provided by the evaluation benchmarks (Table~\ref{tab:quant} in the main paper) are sparse or low-resolution (\eg, LiDAR depth in KITTI~\cite{geiger2013vision} and nuScenes~\cite{caesar2020nuscenes} from SpatialRGPT), which introduce a significant gap compared with the dense point-map inputs used during training. To mitigate this issue, we adopt Prior Depth Anything~\cite{wang2025depthprior} to densify the raw GT depth maps, and sending the refined point maps derived from them as input to our RGB-D VLM for evaluation.

\subsection{More Training Details}
In this section, we provide additional training details, including the initialization strategy, other hyper-parameters and computation details, and the task sampling ratios used during training.

\paravspace
\paragraph{Model initialization.} The Conv2D layer in our 3D branch is initialized via Kaiming initialization~\cite{he2015delving}. 
The linear projector can be naturally split into two parts, for RGB features and point map features respectively. The weights for the RGB part are inherited from the pretrained PaliGemma-2 model, while those for the point map are initialized to zero to ensure that introducing point map at the start of training does not negatively affect the model’s performance.

\begin{table*}[t]
\small
\centering
\caption{Task statistics of our training dataset.}
\begin{tabular}{l l r r}

\toprule
\textbf{Task} & \textbf{Level} & \textbf{\# of QAs} & \textbf{Sampling Ratio in Training} \\
\midrule
Pixel-Wise 3D Point Querying & 0 & Online Generated & 10.51\% \\
Pairwise Depth Ordering & 0 & Online Generated & 2.69\% \\
\midrule
Object Orientation & 1 & 238,909,307 & 3.51\% \\
Object Size & 1 & 55,187,863 & 11.56\% \\
Object Localization & 1 & 36,469,102 & 8.31\% \\
\midrule
Relative Direction & 2 & 510,297,825 & 26.09\% \\
Relative Distance & 2 & 344,264,312 & 13.49\% \\
Relational Comparison & 2 & 507,908,756 & 11.53\% \\
\midrule
Perspective Taking & 3 & 709,415,000 & 4.22\% \\
Spatial Object Counting & 3 & 35,893,426 & 7.52\% \\
Spatial Problem Solving & 3 & 49,322 & 0.57\% \\
\bottomrule
\end{tabular}
\label{tab:tasks}
\end{table*}

\paravspace
\paragraph{Hyper-parameters and computation.} During training, we use AdamW optimizer with a learning rate of $2\times10^{-5}$ and a weight decay of $0.1$. The model is trained on 32 NVIDIA H100 GPUs for approximately 2 days.

\paravspace
\paragraph{Task sampling ratios.} We re-balance tasks from different levels to maintain a reasonable proportion among them during training. The detailed sampling ratios of different tasks are presented in Table~\ref{tab:tasks}.

\subsection{Ablation Study Details}

\paragraph{Inter-level task dependencies.}
In this ablation, we remove certain tasks during training as described in the main paper. To ensure a fair comparison, we maintain the default data sampling distribution. Specifically, whenever a data sample belonging to an excluded task level is drawn from the dataloader, we replace it with a general VQA sample. This replacement strategy ensures that the absolute sampling frequencies and relative proportions of all remaining tasks remain strictly identical to the default configuration.

\paragraph{Relative depth input.} For the variant with relative depth input as discussed in Sec.~\ref{sec:ablation} in the main paper, we retain the original network architecture but only modify its 3D input branch. Specifically, we replace the 3D point coordinates in the metric point map with three identical copies of the relative depth value. Each depth value is normalized and discretized to an integer in the range $[0, 255]$, following a common normalization convention used in previous approaches~\cite{cheng2024spatialrgpt,daxberger2025mm,liu2025ssr,chen2025sd,zhou2025roborefer}. All other parts and training settings remain identical to the default configuration.

\subsection{Evaluation Details}
\paragraph{Benchmarks}
We evaluate our model on a diverse set of spatial understanding benchmarks. For qualitative tasks, including the full sets of EmbSpatial~\cite{du2024embspatial} and 3DSRBench~\cite{ma20253dsrbench}, as well as the configuration subset of RoboSpatial-Home~\cite{song2025robospatial}, we report Accuracy since these are framed as multiple-choice questions. For quantitative tasks involving numerical estimation, we report the Success Rate based on relative error thresholds: following SpatialRGPT~\cite{cheng2024spatialrgpt}, a prediction is successful if the relative error is within 25\%; for Q-Spatial, we adopt its original criterion where a relative error within 50\% is considered a success.

\paragraph{Ablation Study}
In the ablation study, we select subsets of 3DSRBench to report Level 2 (L2) and Level 3 (L3) performance:

\begin{itemize}
    \RaggedRight 
    \item \textbf{3DSR-L2}: height\_higher, location\_above, location\_closer\_to\_camera, location\_next\_to, multi\_object\_closer\_to, multi\_object\_facing, multi\_object\_parallel, multi\_object\_same\_direction.
    \item \textbf{3DSR-L3}: orientation\_in\_front\_of, orientation\_on\_the\_left, multi\_object\_viewpoint\_towards\_object.
\end{itemize}

\section{Dataset Construction Details}
\subsection{Web Data Preprocessing}\label{sec:web_data}

\paragraph{Image filtering.}
We begin with a collection of 15M web images from KosMos-2~\cite{peng2023kosmos} and apply a series of filtering steps to remove non-natural photographs (\eg, charts, forms, GUIs, code snippets):

\vspace{4pt}
\noindent\textit{(1) CLIP-based semantic filtering.} Following SpatialVLM, we employ CLIP~\cite{radford2021learning} for semantic filtering. We construct two tag sets: an \emph{include} set containing text descriptions indicating that an image is a natural photograph, and an \emph{exclude} set containing those indicating that an image is non-natural. All text descriptions in both sets are encoded using the CLIP text encoder. For each image, we obtain its embedding via the CLIP image encoder and retrieve the top-$5$ text tags with the highest similarity. An image is retained if more than half of its retrieved tags belong to the \emph{include} set.

\vspace{4pt}
\noindent\textit{(2) Heuristic filtering.}  We then remove low-quality or non-visual images using simple pixel-based heuristics. Specifically, we discard images in which more than 35\% of the pixels are pure white or pure black, as well as images for which over 50\% of pixels have invalid depth estimates, according to MoGe-2’s validity masks. These empirical rules effectively filter out remaining GUI, chart, table, or blueprint images.

\vspace{4pt}
\noindent\textit{(3) VLM-based filtering.} Finally, we adopt an VLM to refine the filtering results. Following RoboRefer, we use Qwen2.5-VL with the same prompt template to identify and remove images that still lack meaningful spatial information.

\paravspace
\paragraph{Spatial information estimation}
We employ several specialized models to extract object-level point clouds. Specifically, we first use MoGe-2 to estimate both the metric point map and camera intrinsics. Then, following SpatialRGPT, and as described in Sec.~\ref{sec:data_pipe} in the main paper, we compute the point clouds for all detected objects in the image. 

\begin{figure*}[p]
    \centering
    \includegraphics[width=\textwidth]{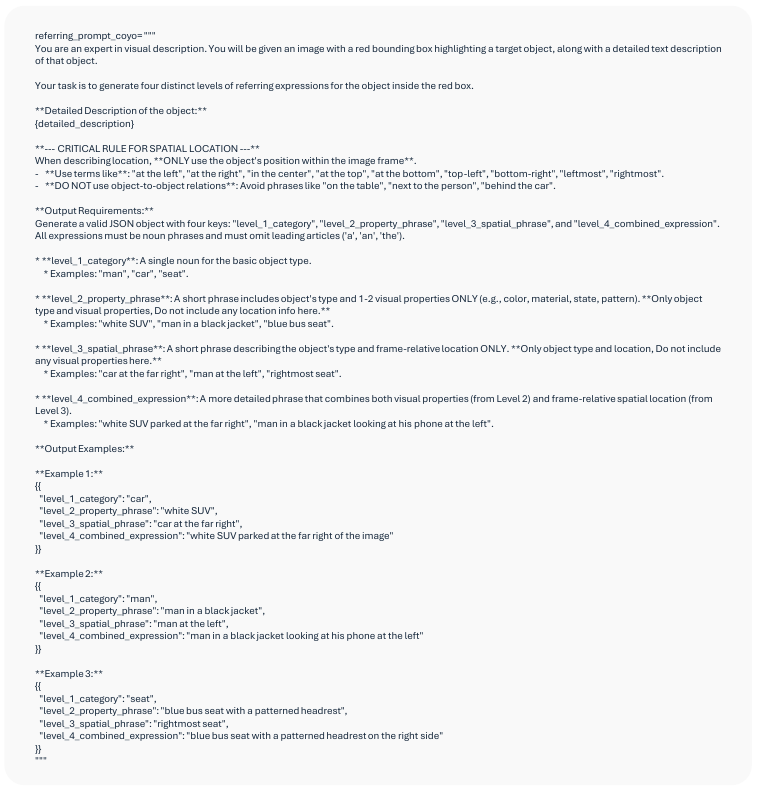}
    \caption{Prompt used to generate the object reference in kosmos dataset}
    \label{fig:referring_prompt_coyo}
    \vspace{-4pt}
\end{figure*}

After obtaining the raw object-level point clouds, we apply DBSCAN clustering~\cite{ester1996density} to identify objects containing multiple point cloud clusters (caused by 2D occlusions during segmentation or cases where multiple objects are grouped into a single segmentation). We select the largest point cloud cluster within each object. 
\paravspace
\paragraph{Textual reference generation.}\label{sec:text_refer}
We first use the Describe Anything model to produce high-quality, detailed captions for each object. Then, following a structured protocol, we use Qwen2.5-VL to generate four levels of object references conditioned on both the image and its corresponding detailed caption. The detailed prompt is shown in Fig.~\ref{fig:referring_prompt_coyo}.

\begin{figure*}[p]
    \centering
    \includegraphics[width=\textwidth]{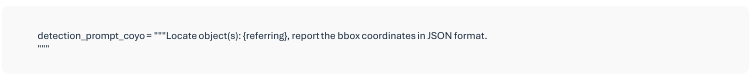}
    \caption{Prompt used to vlm grounding for evaluation in kosmos dataset}
    \label{fig:detection_prompt_coyo}
    \vspace{-4pt}
\end{figure*}

\vspace{6pt}
For each generated reference, we verify its uniqueness using a VLM-based grounding procedure. We prompt Qwen2.5-VL to localize the referred object in the original image. If the model predicts multiple bounding boxes, the reference is discarded, as it may correspond to more than one object. If exactly one bounding box is returned, we further extract the object mask using SAM and compute its IoU with the ground-truth mask. References with IoU greater than 0.7 are retained as valid. The detailed prompt is shown in Fig.~\ref{fig:detection_prompt_coyo}.
Finally, for each object, we select the simplest reference level that passes verification. If no generated reference passes, we fall back to a bounding-box-based reference pattern, such as: ``the [object class] highlighted by the [bbox color] box''.

\subsection{Objects365 Data Preprocessing}\label{sec:o365_data}

\paragraph{Image filtering.}
Objects365~\cite{shao2019objects365} consists of natural photographs collected from Flickr. Therefore, unlike web data, we do not apply additional image filtering.

\paravspace
\paragraph{Spatial information estimation.}
We follow the same spatial information estimation pipeline as described in Sec.~\ref{sec:web_data}.

\begin{figure*}[p]
    \centering
    \includegraphics[width=\textwidth]{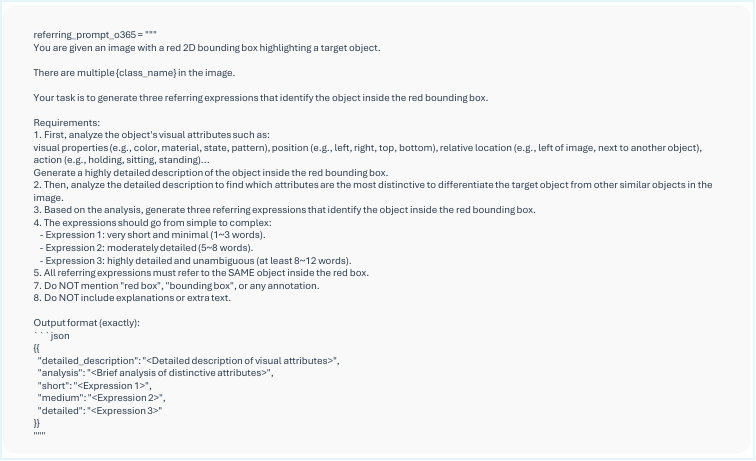}
    \caption{Prompt used to generate the object reference in objects365 dataset}
    \label{fig:referring_prompt_o365}
    \vspace{-4pt}
\end{figure*}

\begin{figure*}[p]
    \centering
    \includegraphics[width=\textwidth]{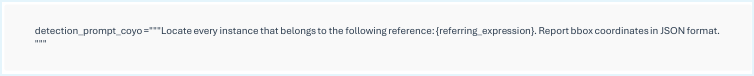}
    \caption{Prompt used to vlm grounding for evaluation in objects365 dataset}
    \label{fig:detection_prompt_o365}
    \vspace{-4pt}
\end{figure*}

\paravspace
\paragraph{Textual reference generation.}
For each annotated object, we directly use Qwen3-VL to generate referring expressions. The prompt template is shown in Fig.~\ref{fig:referring_prompt_o365}. We use the same verification procedure as described in Sec.~\ref{sec:text_refer} to ensure that each reference uniquely identifies a single object. The verification prompt is shown in Fig.~\ref{fig:detection_prompt_o365}.

\paravspace
\paragraph{Task-oriented QA synthesis.}
Compared with web data, Objects365 provides object category labels and 2D bounding box annotations. Based on these annotations, we additionally introduce a \emph{spatial object counting} task.

\subsection{CA-1M Data Preprocessing}
\label{sec:ca1m_pipe}

\paragraph{Category labeling.}

Although each frame in the CA-1M dataset is annotated with relatively complete 3D bounding boxes, these 3D boxes generally lack category-level labels. In most cases, the category field is simply marked as “object” rather than a specific semantic class. Therefore, it is necessary to assign semantic category labels to these bounding boxes in order to support the subsequent generation of object references.

To begin with, we generate 2D bounding box predictions with explicit semantic categories for frames from CA-1M. Following the data pipeline described in Sec.~3.2 in the main paper, we use RAM to recognized object categories in the images and GroundingDINO to obtain corresponding 2D bounding boxes. Additionally, we leverage Qwen2.5-VL to provide a more comprehensive set of categories for GroundingDINO, complementing RAM’s predictions\footnote{We observe that RAM occasionally produces overly coarse labels for indoor objects, \eg, assigning the generic term “furniture”.}.

We then match these category-aware 2D bounding box predictions with GT bounding box annotations provided by CA-1M. We compute the IoU-based cost between boxes and apply Hungarian matching to identify the optimal assignment. To ensure reliable correspondences, any matched pair whose IoU is below 0.4 is discarded. Finally, the semantic category of each matched predicted bounding box is assigned to its corresponding ground-truth box.

\paravspace
\paragraph{Spatial reference generation.} 
Relying solely on category labels is insufficient for accurately referring to objects in downstream QA tasks, especially when multiple instances of the same category are present in a scene. Therefore, we further incorporate spatial references to disambiguate objects, complementing the textual references introduced in Sec.~\ref{sec:text_refer}. When several objects share the same category label, we distinguish among them using their spatial position relationships or their size properties, such as “the second farthest chair” or “the largest pillow”. More specifically, the spatial references can be categorized into the following types:

\vspace{4pt}
\noindent\textit{(1) Category-based reference.} 
When there is only a single instance of a given category in the scene, the object can be directly referred to using its category label alone, such as ``the remote control'', ``the sofa'', and ``the television.''

\vspace{4pt}
\noindent\textit{(2) Global intra-category relations.} 
When multiple objects of the same category are present, we generate references based on their global positional or size relationships. These relations primarily include the following three forms:

\begin{itemize}
\setlength{\itemindent}{8pt}
    \item \emph{Linear ordering.} If all instances of the same category exhibit a roughly linear spatial arrangement, the object can be referred to by its ordinal position along that line. 
To detect such linear distributions, we apply PCA to the center positions of all objects in the category. 
If the second principal component is smaller than $15\%$ of the first principal component, we treat the objects as lying along the dominant principal axis. 
The orientation of this axis determines whether the linear ordering corresponds to a ``left--right,'' ``front--back,'' or ``top--bottom'' direction. An example is ``the second picture frame from the left''.

\vspace{2pt}
\item \emph{Positional comparison.} We generate spatial references by comparing the relative positions of objects within the same category along canonical spatial axes (\eg, front, back, left, right, top, bottom, near, far). 
Such comparisons allow us to uniquely identify an object based on its extremal or ranked position in space. For example, ``the leftmost bowl'' and ``the second closest table to the camera''.

\vspace{2pt}
\item \emph{Size comparison.} Since the objects in CA-1M are annotated with accurate length, width, and height measurements, we can distinguish among instances of the same category by comparing their size properties, such as ``the widest sofa'' and ``the largest door''.

\end{itemize}

\vspace{4pt}
It is worth noting that the accuracy of spatial references depends on both the recall of all instances in the scene and the completeness of their category annotations, which is facilitated by the detailed spatial bounding box annotations in CA-1M dataset and further enhanced by our category labeling procedure which assigns a reliable semantic category to each bounding box.

\paravspace
\paragraph{Task-oriented QA synthesis.}

\begin{figure*}[p]
    \centering
    \includegraphics[width=\textwidth]{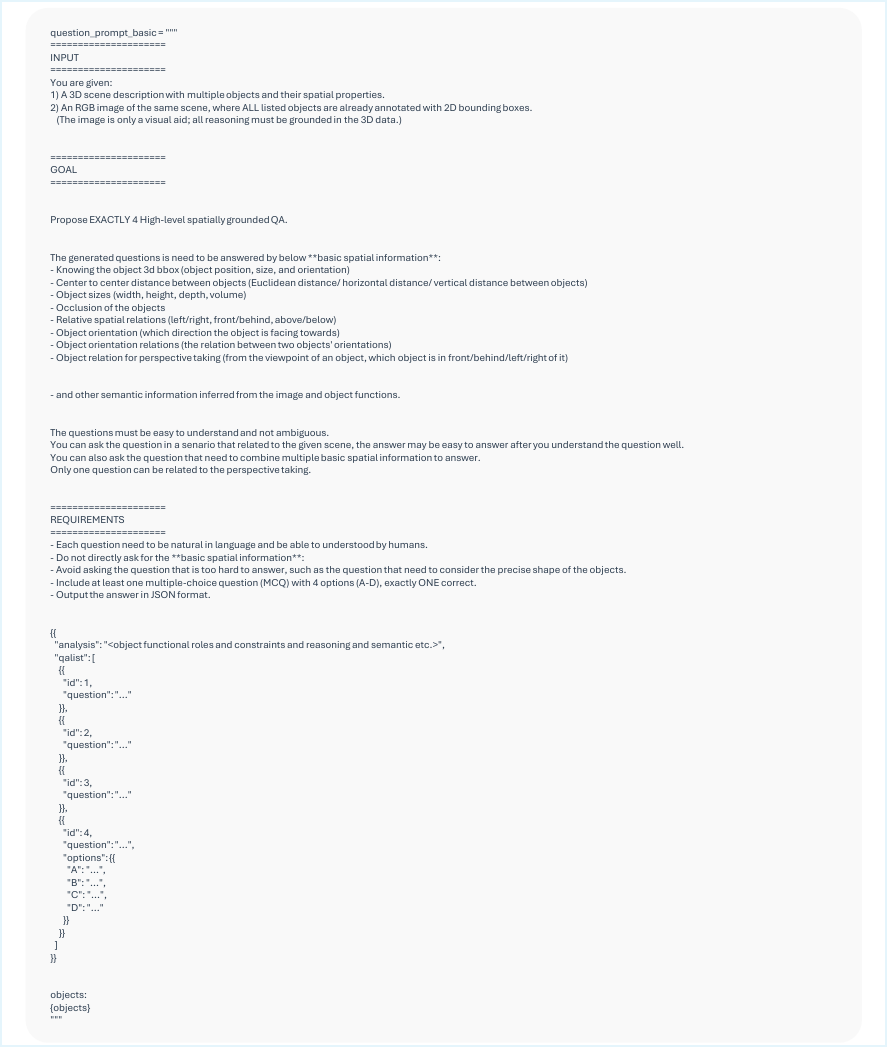}
    \caption{Prompt used to generate the basic problem solving question}
    \label{fig:question_prompt_basic}
    \vspace{-4pt}
\end{figure*}

\begin{figure*}[p]
    \centering
    \includegraphics[width=\textwidth]{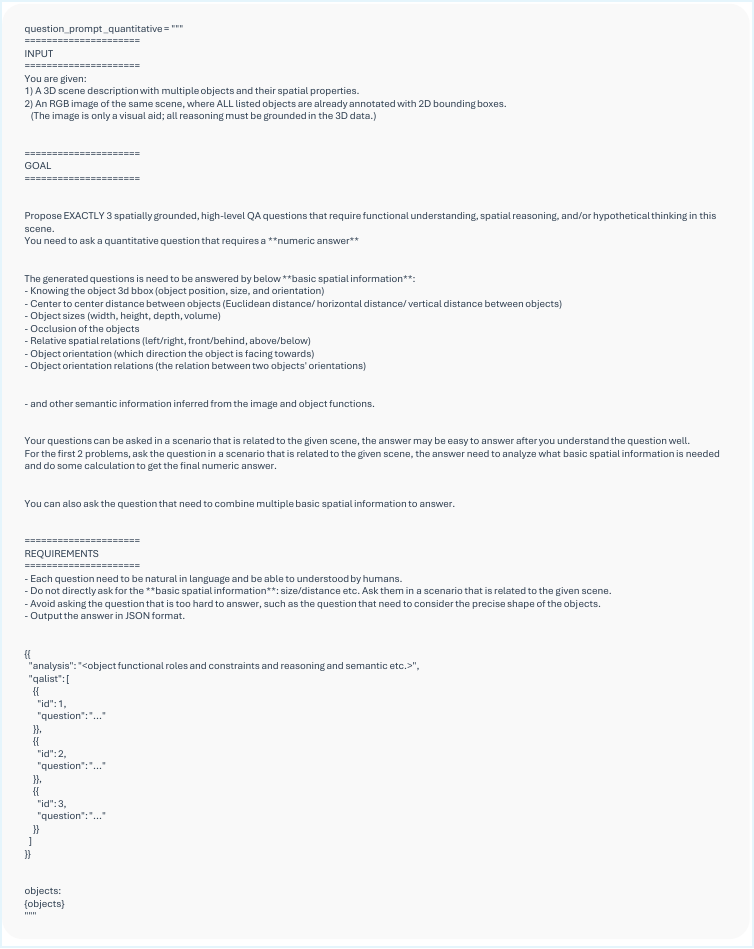}
    \caption{Prompt used to generate the quantative problem solving question}
    \label{fig:question_prompt_quantative}
    \vspace{-4pt}
\end{figure*}

\begin{figure*}[p]
    \centering
    \includegraphics[width=\textwidth]{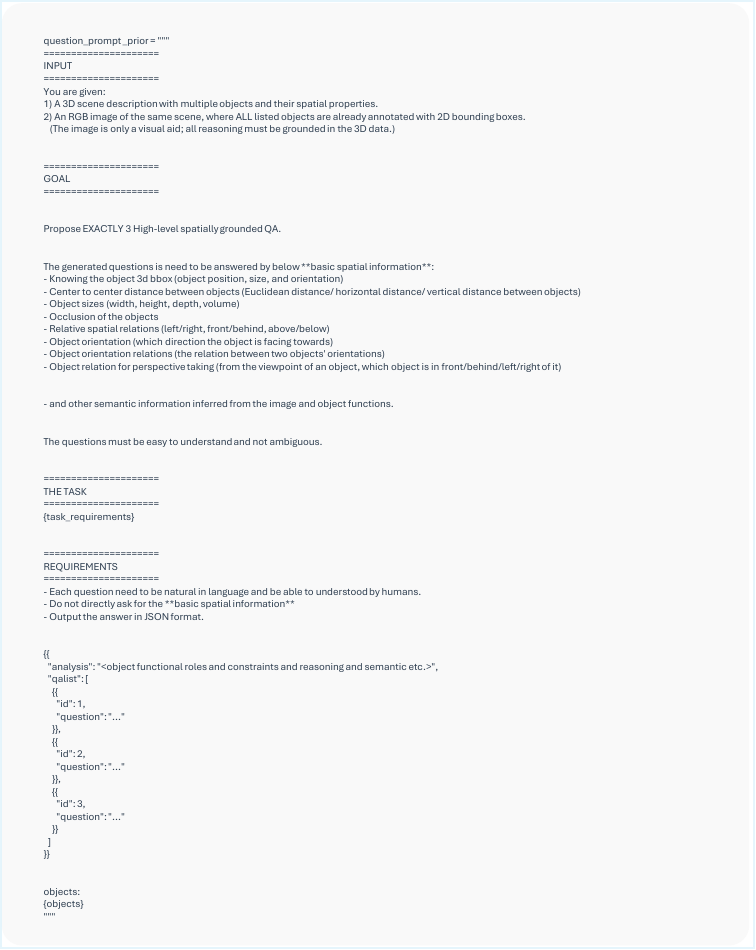}
    \caption{Prompt used to generate the problem solving question guided by few shot demonstrations}
    \label{fig:question_prompt_prior}
    \vspace{-4pt}
\end{figure*}

As described in the main paper, we use GPT-4.1 to generate QA pairs for Level-3 spatial problem-solving tasks. We construct spatial reasoning QA pairs using a multi-stage prompting pipeline grounded in structured 3D scene descriptions. Given the 3D object annotations and the corresponding RGB image, we first generate spatially grounded questions using three prompt templates: (1) basic spatial reasoning questions (Fig.~\ref{fig:question_prompt_basic}), (2) quantitative reasoning questions requiring numerical answers (Fig.~\ref{fig:question_prompt_quantative}), and (3) prior-guided questions generated with few-shot demonstrations (Fig.~\ref{fig:question_prompt_prior}). These prompts encourage questions that require reasoning over object positions, sizes, orientations, distances, and occlusion relations.

\begin{figure*}[p]
    \centering
    \includegraphics[width=\textwidth]{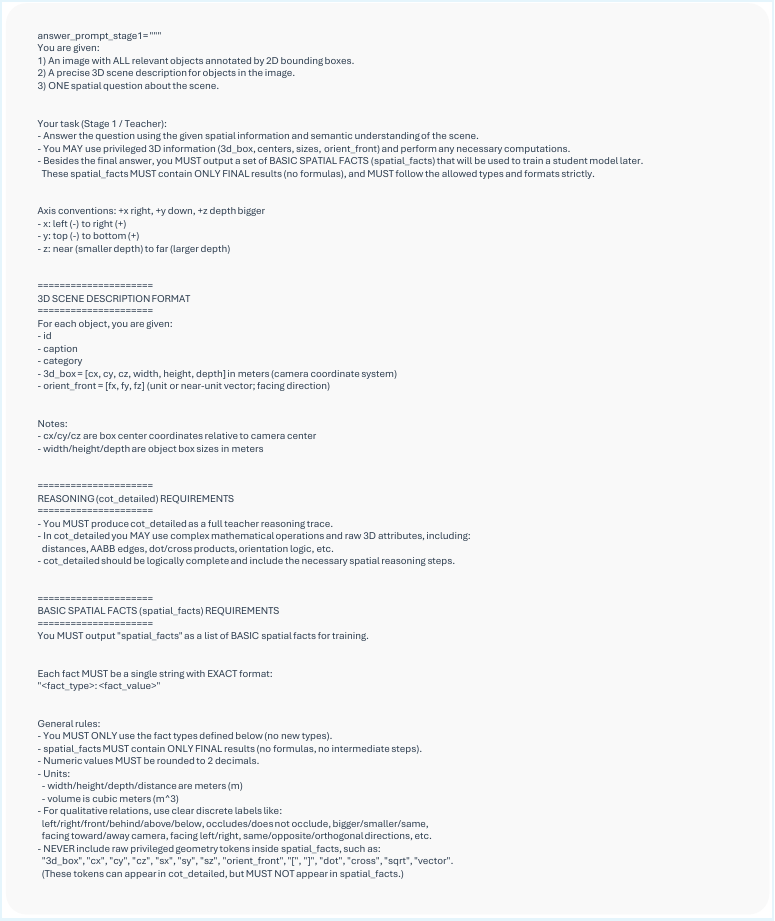}
    \caption{Prompt used to generate the answer of the given problem solving question (stage1, part1)}
    \label{fig:answer_prompt_stage1}
    \vspace{-4pt}
\end{figure*}

\begin{figure*}[p]
    \centering
    \includegraphics[width=\textwidth]{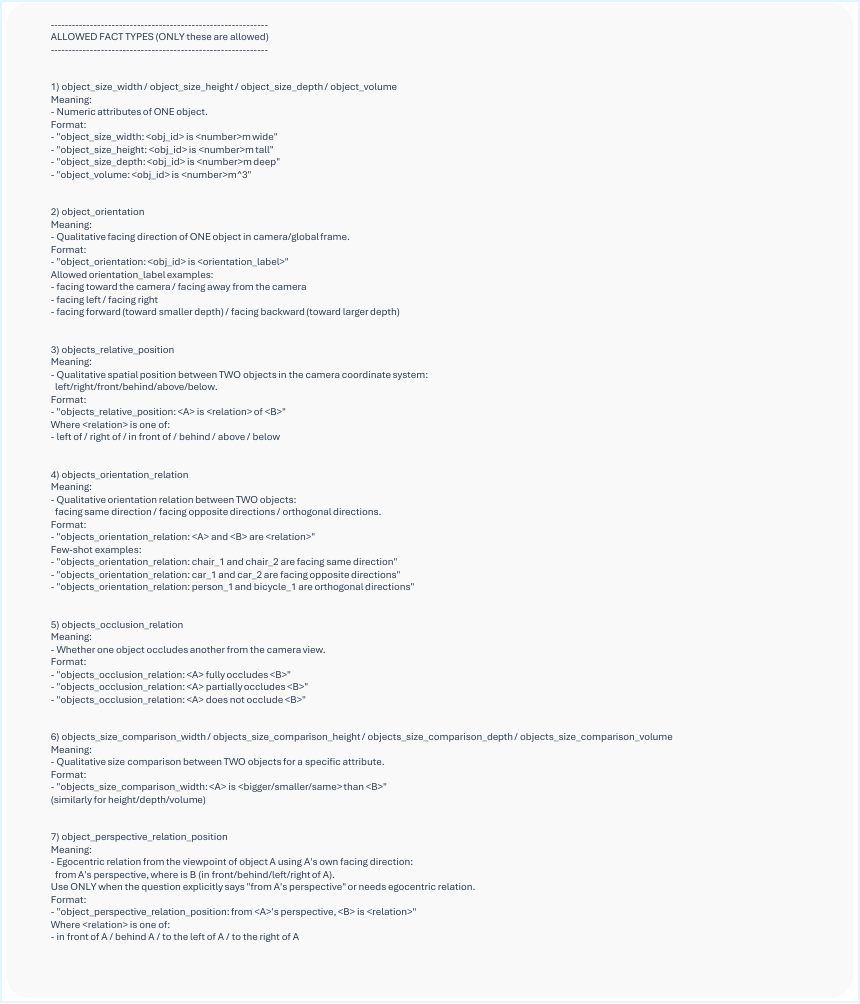}
    \caption{Prompt used to generate the answer of the given problem solving question (stage1, part2)}
    \label{fig:answer_prompt_stage1_2}
    \vspace{-4pt}
\end{figure*}

\begin{figure*}[p]
    \centering
    \includegraphics[width=\textwidth]{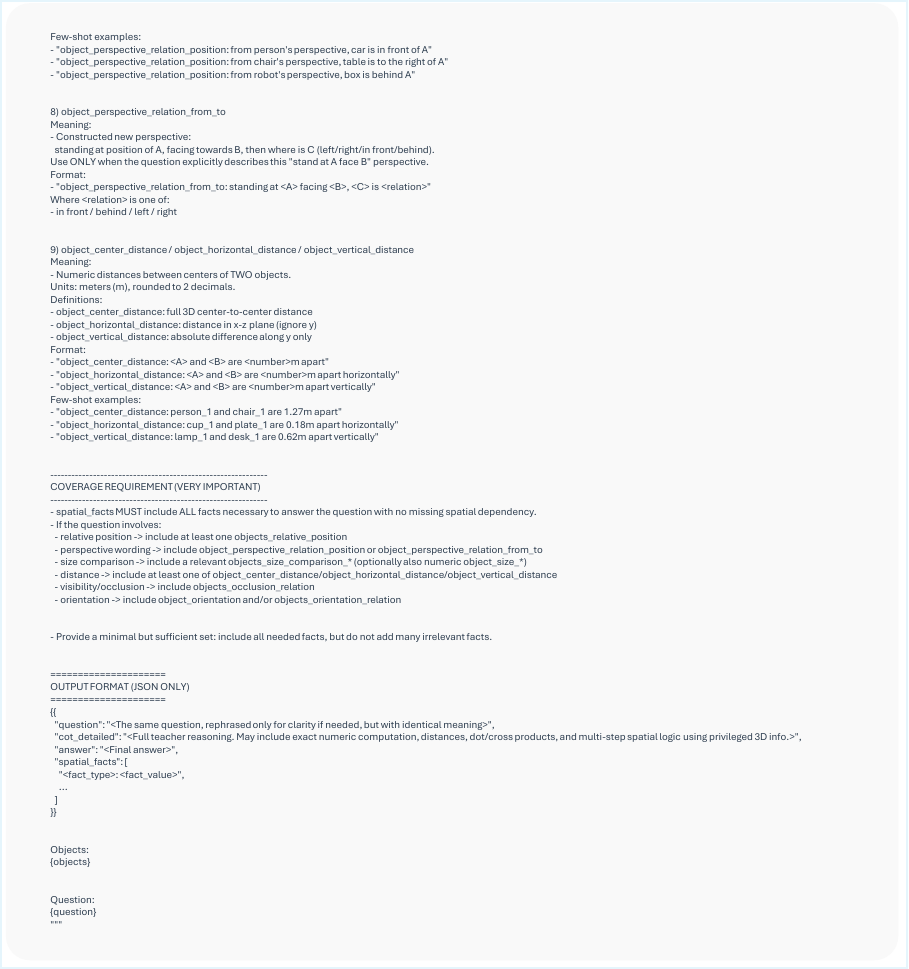}
    \caption{Prompt used to generate the answer of the given problem solving question (stage1, part3)}
    \label{fig:answer_prompt_stage1_3}
    \vspace{-4pt}
\end{figure*}

\begin{figure*}[p]
    \centering
    \includegraphics[width=\textwidth]{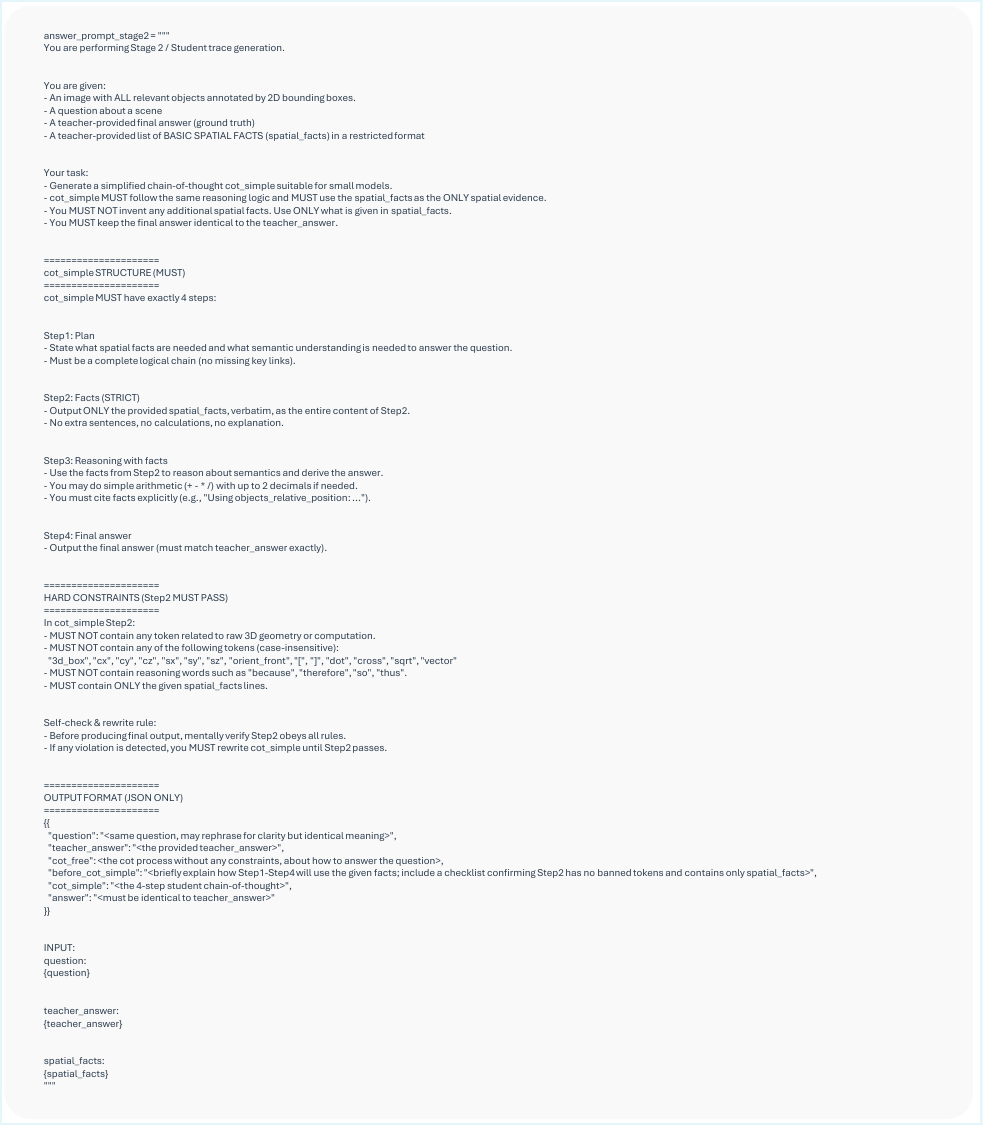}
    \caption{Prompt used to generate the answer of the given problem solving question (stage2)}
    \label{fig:answer_prompt_stage2}
    \vspace{-4pt}
\end{figure*}

We then produce answers and reasoning traces through a two-stage prompting process. In Stage~1, a teacher model answers each question using privileged 3D information and outputs both a detailed reasoning trace and a set of structured \textit{spatial facts} summarizing the key spatial relations required for solving the question (Fig.~\ref{fig:answer_prompt_stage1},~\ref{fig:answer_prompt_stage1_2} and~\ref{fig:answer_prompt_stage1_3}). In Stage~2, the model is prompted to generate a simplified student reasoning trace that derives the answer solely from the provided spatial facts, without access to raw 3D geometry (Fig.~\ref{fig:answer_prompt_stage2}). This pipeline produces QA pairs with structured spatial evidence and aligned reasoning traces for training spatial reasoning models.

\subsection{General VQA Data Preprocessing}
For the general VQA datasets from LLaVA-Next (Sec.~\ref{sec:experiments} in the main paper), we adopt a simplified processing pipeline. Since these datasets already provide well-structured subsets, we do not apply the CLIP-based filtering used in our web-data pipeline (Sec.~\ref{sec:web_data}). Instead, we remove subsets containing non–real-world images, such as synthetic images, diagrams, OCR-focused content, or charts and tables, retaining only those consisting of real-world images. We then apply MoGe-2 to each remaining image to generate the corresponding point map as auxiliary 3D input.

\section{Custom Benchmark} In this section, we provide more details of our custom benchmark described in Sec.~\ref{sec:benchmarks} in the main paper. The benchmark contains three different types of tasks as described below.

\paravspace
\paragraph{Object-to-camera distance \& relative direction estimation.} We constructed these two types of tasks using Omni3D~\cite{brazil2023omni3d}, which provides ground-truth point-cloud information and 3D object bounding boxes. The tasks are built upon Omni3D subsets sourced from KITTI, nuScenes, SUN RGB-D~\cite{song2015sun}, and Hypersim~\cite{roberts2021hypersim}, covering indoor, outdoor, and synthetic scenarios.
We sample images from these sources, and apply the same pipeline as described in Sec.~\ref{sec:web_data} to generate textual references.

After generating object references, we create QA pairs using task-specific templates described in Sec.~\ref{sec:data_pipe} in the main paper. For the \emph{distance estimation} task, every object involved in the question is required to have a unique textual reference. For the \emph{relative direction estimation} task, at least one object mentioned in the question must have a unique textual reference, while other objects are referred to by their textual descriptions accompanied by their corresponding 2D bounding boxes.

In addition, we conduct a manual verification step to ensure the correctness of both references and QA pairs. We remove incorrect references as well as ambiguous QA examples (\eg, directional relations between a sofa and the pillow placed on it).

\begin{figure*}[p]
    \centering
    \includegraphics[width=\textwidth]{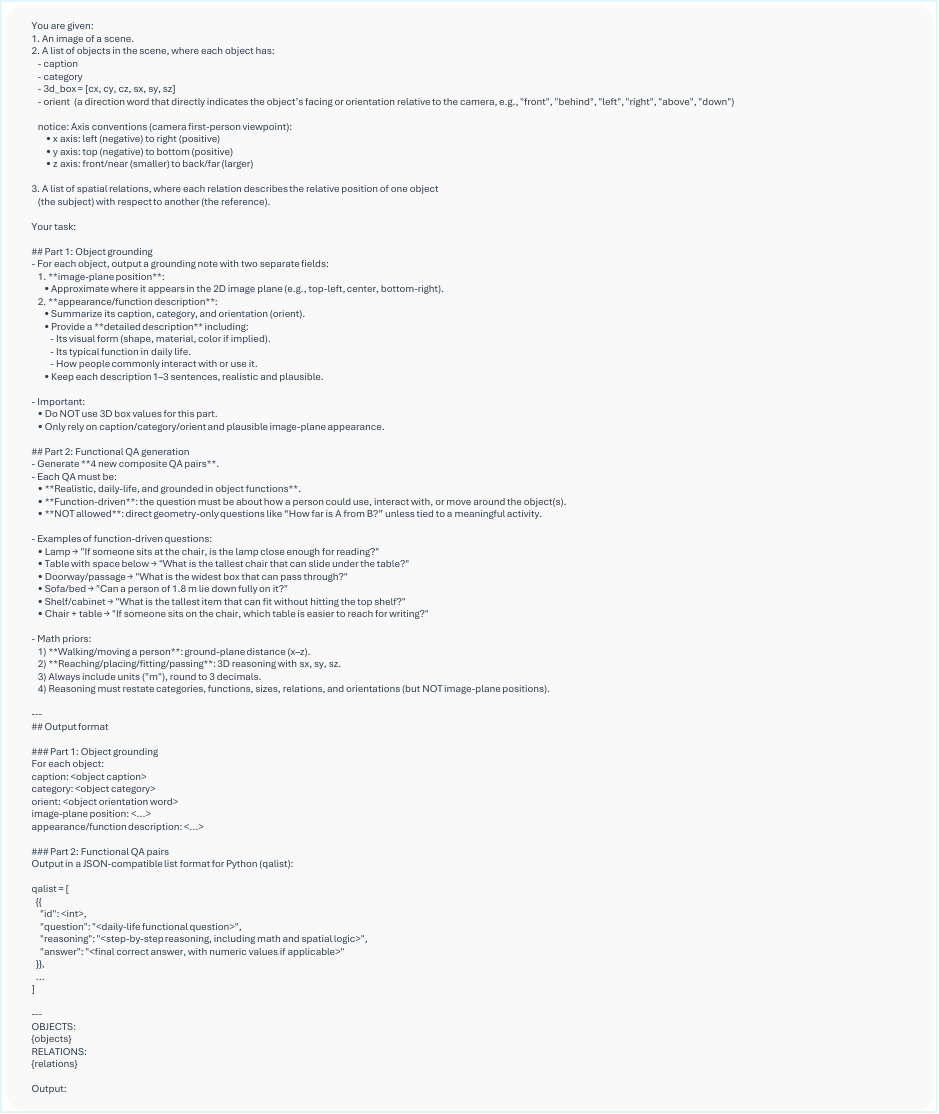}
    \caption{Prompt used to generate Level-3 spatial problem-solving QA pairs.(test set)}
    \label{fig:prompt_solving}
    \vspace{-4pt}
\end{figure*}

\paravspace
\paragraph{Spatial problem solving.} We use the prompt in Fig.~\ref{fig:prompt_solving} to generate problem-solving data from the \emph{validation split} of the CA-1M dataset. The generated QA pairs are then manually inspected to verify their correctness, and inaccurate or ambiguous cases are filtered out.

\begin{figure*}[p]
    \centering
    \includegraphics[width=\textwidth]{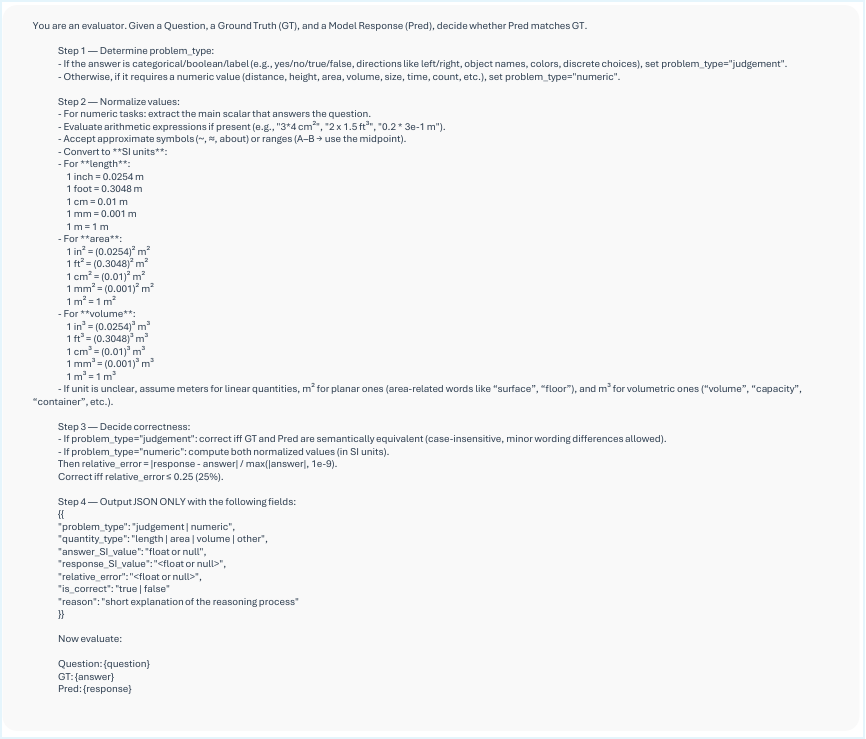}
    \caption{Prompt used to evaluate the correctness of answers in our custom benchmark for spatial problem-solving tasks.}
    \label{fig:prompt_selfbench_eval}
    \vspace{-4pt}
\end{figure*}

While the answers to spatial problem-solving tasks may be in free-form, we employ GPT-4.1 as an auxiliary judging agent to more accurately evaluate their correctness, as illustrated in Fig.~\ref{fig:prompt_selfbench_eval}.
For \emph{judgement} questions—such as category identification or yes/no queries—we prompt GPT to determine whether the model's prediction matches the ground-truth answer. 
For \emph{numeric} questions, GPT is instructed to verify whether the numerical prediction falls within $25\%$ relative error of the ground-truth value.

\section{More Results}

\subsection{More Examples of Constructed Data}

\begin{figure*}[p]
    \centering
    \includegraphics[width=\textwidth]{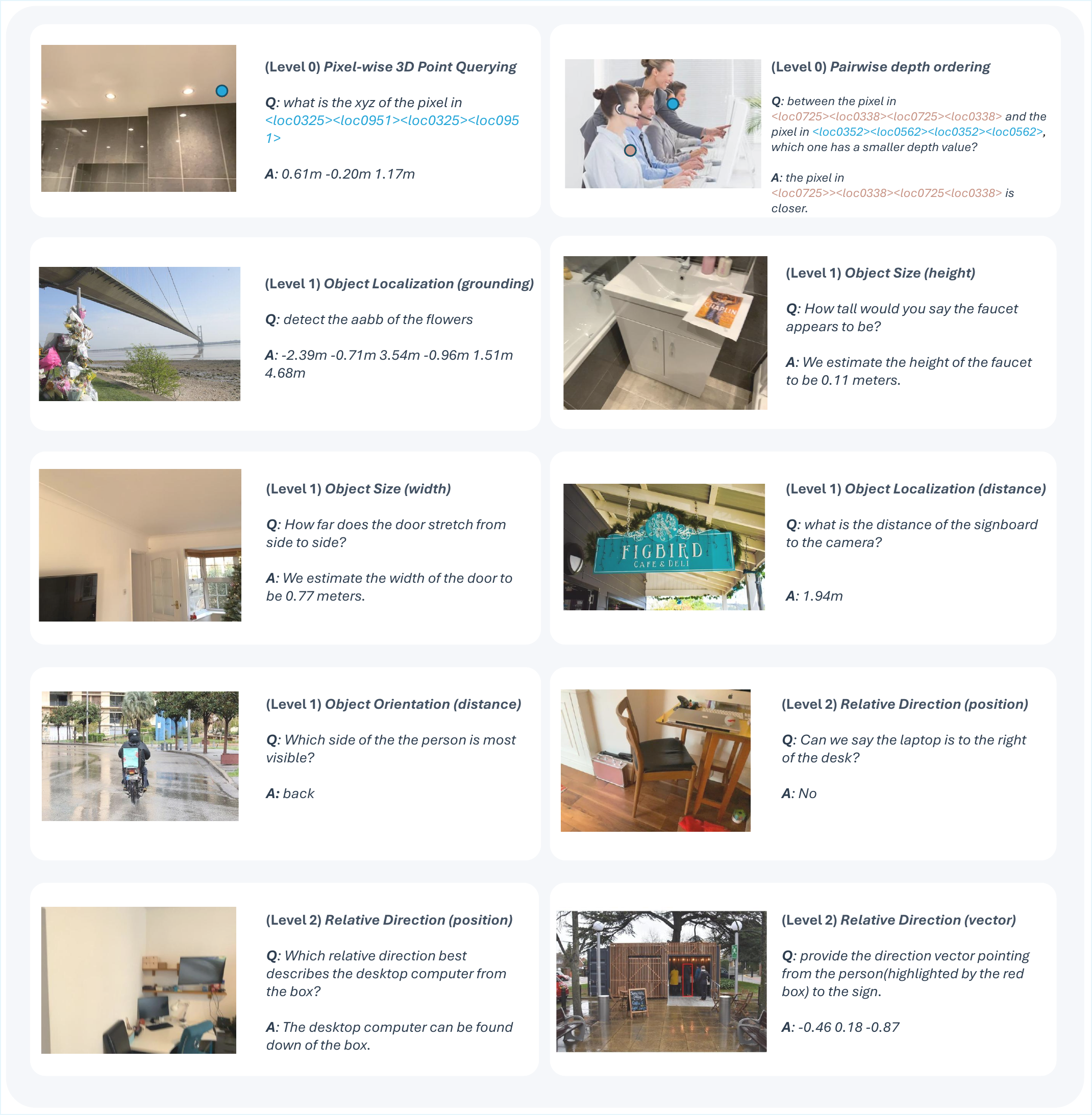}
    \caption{Examples of spatial VQA data constructed using our method, covering different task levels.}
    \label{fig:qa_list1}
    \vspace{-4pt}
\end{figure*}

\begin{figure*}[p]
    \centering
    \includegraphics[width=\textwidth]{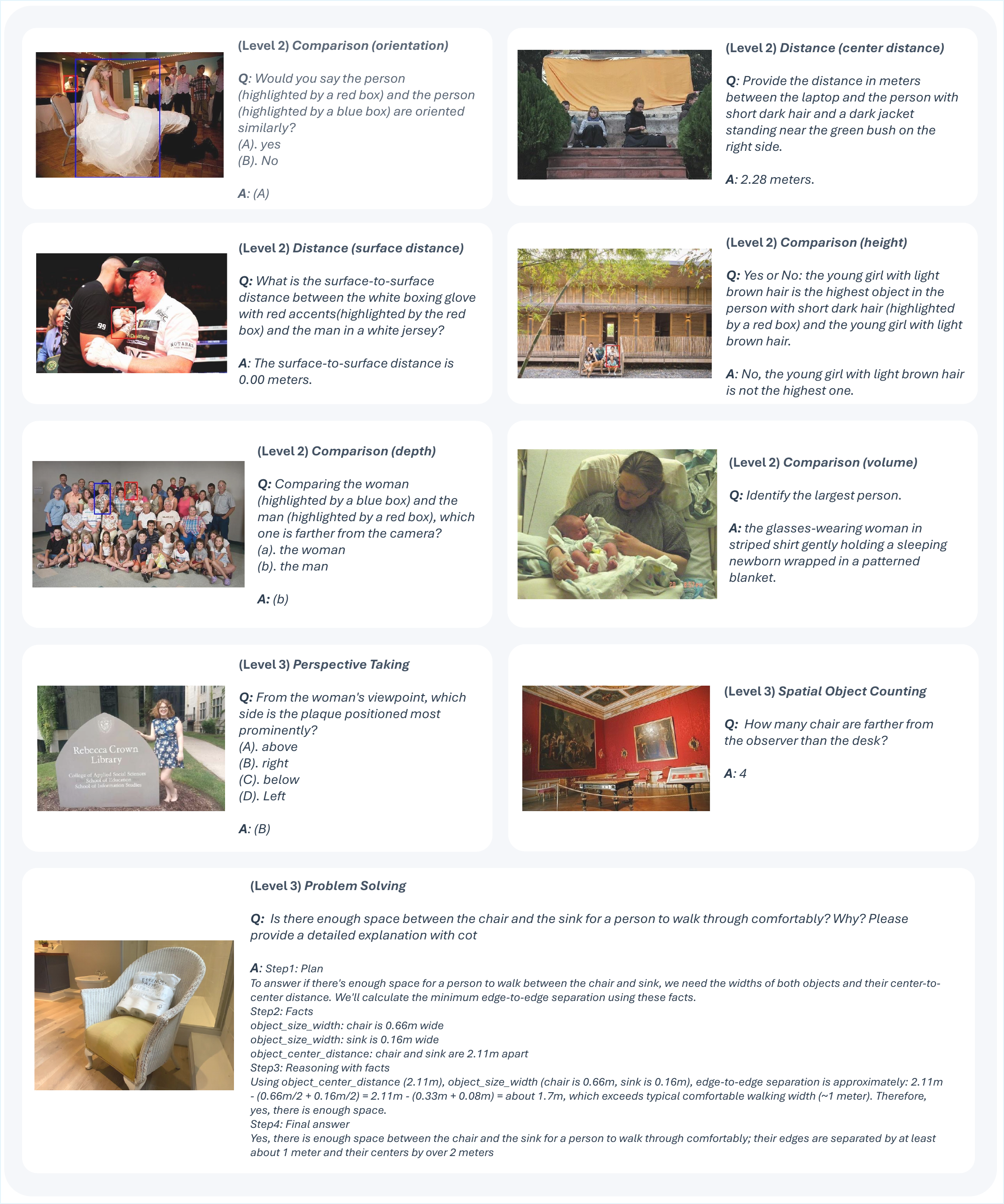}
    \caption{Examples of spatial VQA data constructed using our method, covering different task levels.}
    \label{fig:qa_list2}
    \vspace{-4pt}
\end{figure*}

Figures~\ref{fig:qa_list1} and~\ref{fig:qa_list2} present additional examples of the spatial VQA tasks generated by our method, covering diverse environments and tasks across different levels.

\subsection{More Examples of Model Responses}

\begin{figure*}[p]
    \centering
    \includegraphics[width=\textwidth]{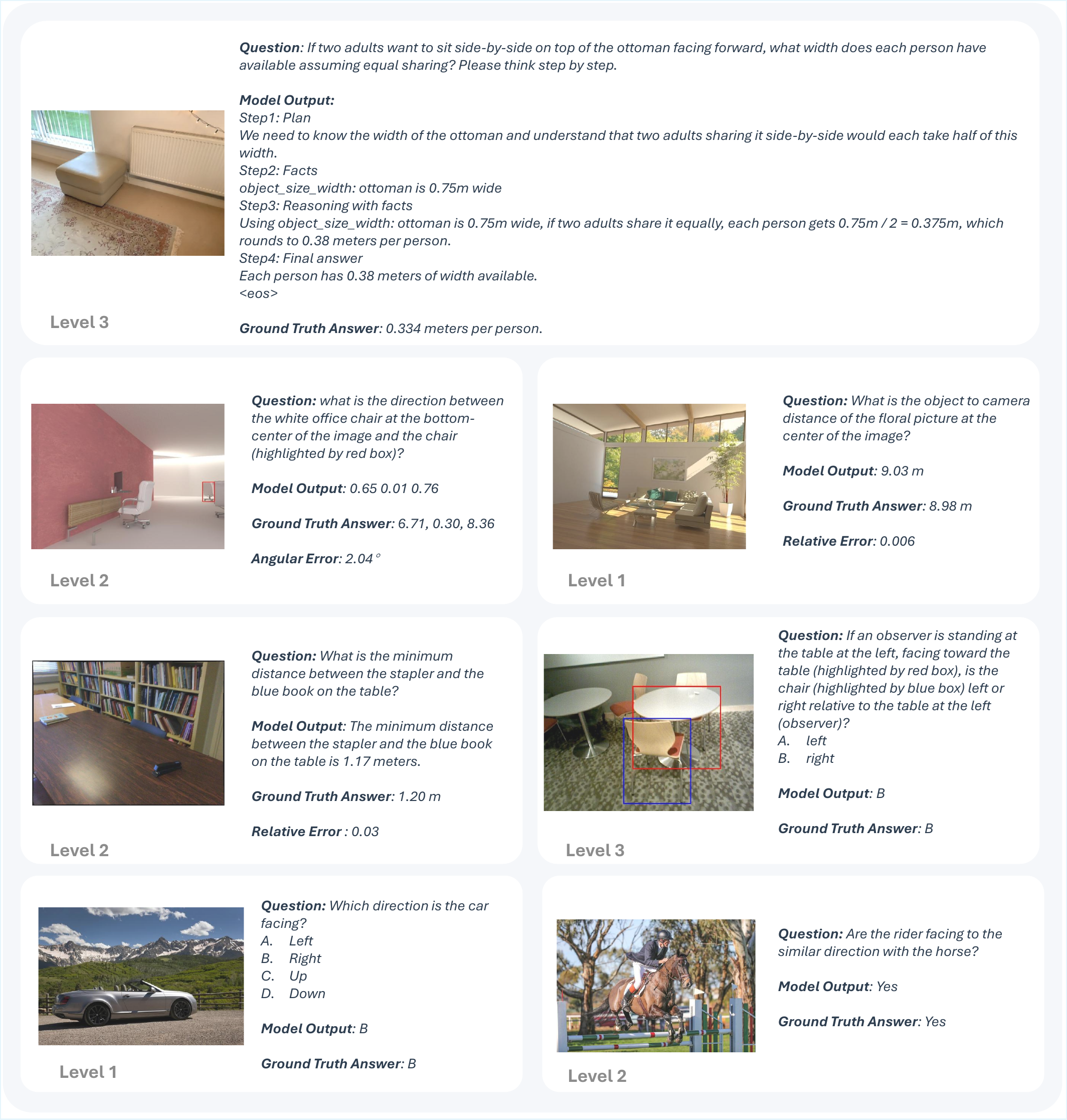}
    \caption{Examples of our model’s responses on unseen images.}
    \label{fig:output}
    \vspace{-4pt}
\end{figure*}

Figure~\ref{fig:output} shows additional test results of our RGB-D VLM on unseen in-the-wild images. Our model is capable of handling tasks at different levels related to 3D spatial understanding and reasoning, producing reasonable and accurate answers.

\section{Limitations and Future Work}
First, the model's generalization capability is currently constrained by both task complexity and linguistic diversity. On the one hand, the model primarily focuses on relatively basic spatial understanding tasks; although we include abstract spatial reasoning at Level 3, it does not comprehensively cover all types of complex reasoning. On the other hand, the procedural nature of our data generation may lead to a reliance on fixed instruction patterns. Consequently, while the model performs well on template-consistent data, its robustness in handling the highly diverse and informal language found in real-world scenarios remains to be further improved. In future work, we plan to introduce a broader set of tasks with natural language variations and incorporate reinforcement fine-tuning (RFT) to enhance the model’s reasoning capabilities on complex spatial problems.

Second, although we highlight the correlations between tasks at different levels in this paper, providing valuable insights for the design of spatial understanding tasks, the current analysis serves only as a starting point. The finer-grained interactions between tasks across levels, as well as the impact of different training strategies on inter-level task relationships, require further investigation and experiments to be fully clarified.

Finally, our model currently only supports monocular input. Many spatial reasoning tasks, however, require an understanding of multi-view scenes or temporal dynamics in videos. We leave these directions for future exploration.

\begin{figure*}[p]
    \centering
    \includegraphics[width=\textwidth]{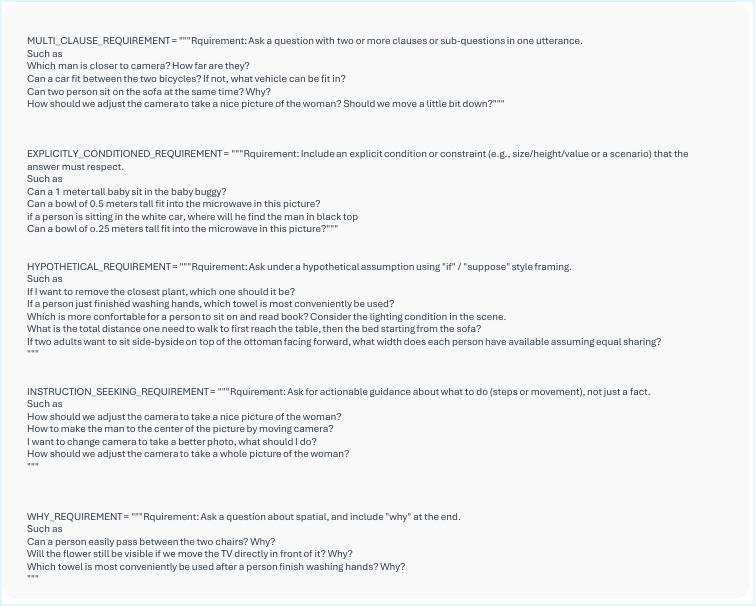}
    \caption{Examples of few-shot demonstration}
    \label{fig:few_shot}
    \vspace{-4pt}
\end{figure*}

